\def\arxiv{true}
\pdfoutput=1

\documentclass[11pt]{article}

\ifx\arxiv\undefined
    \usepackage[review]{ACL2023}
\else
    \usepackage[final]{ACL2023}
\fi

\usepackage{times}
\usepackage[T1]{fontenc}
\usepackage[utf8]{inputenc}
\usepackage{microtype}
\usepackage{inconsolata}

\usepackage{graphicx} 
\usepackage{balance}
\usepackage{amsfonts,amsmath,amsthm,amssymb}
\usepackage{bm}
\usepackage{booktabs} 
\usepackage{threeparttable}
\usepackage[htt]{hyphenat}
\usepackage{enumitem}
\usepackage{xspace}
\usepackage{multirow}
\usepackage{makecell}
\usepackage{hhline}
\usepackage{hyperref}
\usepackage{url}
\usepackage{colortbl}
\usepackage{subfig}
\usepackage[normalem]{ulem} 
\usepackage{algorithm}
\usepackage{appendix}   
\usepackage[noend]{algorithmic}
\usepackage{arydshln}

\usepackage{xspace}
\usepackage{amsmath}
\usepackage{amsfonts}
\usepackage{url}
\usepackage{marvosym}
\usepackage{ifthen}
\theoremstyle{plain}

\newcommand{\chatoDisplayMode}[1]{#1}

\definecolor{MyRed}{rgb}{0.6,0.0,0.0} 
\definecolor{MyBlack}{rgb}{0.1,0.1,0.1} 
\newcommand{\inred}[1]{{\color{MyRed}\sf\textbf{\textsc{#1}}}}
\newcommand{\frameit}[2]{
  \begin{center}
  {\color{MyRed}
  \framebox[.9\columnwidth][l]{
    \begin{minipage}{.85\columnwidth}
    \inred{#1}: {\sf\color{MyBlack}#2}
    \end{minipage}
  }\\
  }
  \end{center}
}

\newcommand{\note}[2][]{\chatoDisplayMode{\def\@tmpsig{#1}\frameit{{\Pointinghand} Note}{#2\ifx \@tmpsig \@empty \else \mbox{ --\em #1}\fi}}}
\newcommand{\todo}[2][]{\chatoDisplayMode{\def\@tmpsig{#1}\frameit{{\Writinghand} To-do}{#2\ifx \@tmpsig \@empty \else \mbox{ --\em #1}\fi}}}

\newcommand{\abbrevStyle}[1]{#1}

\newcommand{\ie}{\abbrevStyle{i.e.}\xspace}
\newcommand{\eg}{\abbrevStyle{e.g.}\xspace}
\newcommand{\cf}{\abbrevStyle{cf.}\xspace}

\newcommand{\vs}{\abbrevStyle{vs.}\xspace}
\newcommand{\etc}{\abbrevStyle{etc.}\xspace}

\newcommand{\xhdr}[1]{\vspace{1.7mm}\noindent{{\bf #1.}}}

\newcommand{\textcite}[1]{\citeauthor{#1} \shortcite{#1}}

\newcommand{\hide}[1]{}

\makeatletter
\newcommand{\iffont}[2]{\ifthenelse{\equal{\f@family}{#1}}{#2}{}}
\makeatother

\iffont{ptm}{
  \usepackage{mathptmx}

  \DeclareSymbolFont{greek}{OML}{cmm}{m}{n}
  \DeclareMathSymbol{\alpha}{\mathalpha}{greek}{"0B}
  \DeclareMathSymbol{\beta}{\mathalpha}{greek}{"0C}
  \DeclareMathSymbol{\gamma}{\mathalpha}{greek}{"0D}
  \DeclareMathSymbol{\delta}{\mathalpha}{greek}{"0E}
  \DeclareMathSymbol{\epsilon}{\mathalpha}{greek}{"0F}
  \DeclareMathSymbol{\zeta}{\mathalpha}{greek}{"10}
  \DeclareMathSymbol{\eta}{\mathalpha}{greek}{"11}
  \DeclareMathSymbol{\theta}{\mathalpha}{greek}{"12}
  \DeclareMathSymbol{\iota}{\mathalpha}{greek}{"13}
  \DeclareMathSymbol{\kappa}{\mathalpha}{greek}{"14}
  \DeclareMathSymbol{\lambda}{\mathalpha}{greek}{"15}
  \DeclareMathSymbol{\mu}{\mathalpha}{greek}{"16}
  \DeclareMathSymbol{\nu}{\mathalpha}{greek}{"17}
  \DeclareMathSymbol{\xi}{\mathalpha}{greek}{"18}
  \DeclareMathSymbol{\pi}{\mathalpha}{greek}{"19}
  \DeclareMathSymbol{\rho}{\mathalpha}{greek}{"1A}
  \DeclareMathSymbol{\sigma}{\mathalpha}{greek}{"1B}
  \DeclareMathSymbol{\tau}{\mathalpha}{greek}{"1C}
  \DeclareMathSymbol{\upsilon}{\mathalpha}{greek}{"1D}
  \DeclareMathSymbol{\phi}{\mathalpha}{greek}{"1E}
  \DeclareMathSymbol{\chi}{\mathalpha}{greek}{"1F}
  \DeclareMathSymbol{\psi}{\mathalpha}{greek}{"20}
  \DeclareMathSymbol{\omega}{\mathalpha}{greek}{"21}
  \DeclareMathSymbol{\varepsilon}{\mathalpha}{greek}{"22}
  \DeclareMathSymbol{\vartheta}{\mathalpha}{greek}{"23}
  \DeclareMathSymbol{\varpi}{\mathalpha}{greek}{"24}
  \DeclareMathSymbol{\varrho}{\mathalpha}{greek}{"25}
  \DeclareMathSymbol{\varsigma}{\mathalpha}{greek}{"26}
  \DeclareMathSymbol{\varphi}{\mathalpha}{greek}{"27}
  \DeclareSymbolFont{otone}{OT1}{cmr}{m}{n}
  \DeclareMathSymbol{\Gamma}{\mathalpha}{otone}{0}
  \DeclareMathSymbol{\Delta}{\mathalpha}{otone}{1}
  \DeclareMathSymbol{\Theta}{\mathalpha}{otone}{2}
  \DeclareMathSymbol{\Lambda}{\mathalpha}{otone}{3}
  \DeclareMathSymbol{\Xi}{\mathalpha}{otone}{4}
  \DeclareMathSymbol{\Pi}{\mathalpha}{otone}{5}
  \DeclareMathSymbol{\Sigma}{\mathalpha}{otone}{6}
  \DeclareMathSymbol{\Upsilon}{\mathalpha}{otone}{7}
  \DeclareMathSymbol{\Phi}{\mathalpha}{otone}{8}
  \DeclareMathSymbol{\Psi}{\mathalpha}{otone}{9}
  \DeclareMathSymbol{\Omega}{\mathalpha}{otone}{10}
  \DeclareSymbolFont{syms}{OML}{cmm}{m}{it}
  \DeclareMathSymbol{\partial}{\mathord}{syms}{"40}
  \DeclareMathAlphabet{\mathbold}{OML}{cmm}{b}{it}
  \DeclareSymbolFont{largesymbols}{OMX}{cmex}{m}{n}

}

\newcommand{\ignore}[1]{}

\newcommand{\tabcaption}[1]{\vspace*{-3mm}\caption{#1}\vspace*{-4mm}}
\newcommand{\figcaption}[1]{\vspace*{-3mm}\caption{#1}\vspace*{-5mm}}
\newcommand{\moveup}{\vspace*{-2mm}}
\newcommand{\moveups}{\vspace*{-1mm}}

\newcommand{\tomas}[1]{\textcolor{black}{#1}}

\newcommand{\loki}{\textsc{LocEI}\xspace}
\newcommand{\xloki}{\textsc{xLocEI}\xspace}
\newcommand{\xlokismall}{\textsc{xLocEI}$_{11}$\xspace}
\newcommand{\xlokilarge}{\textsc{xLocEI}$_{20}$\xspace}

\title{Entity Insertion in Multilingual Linked Corpora: The Case of Wikipedia}

\DeclareSymbolFont{extraup}{U}{zavm}{m}{n}
\DeclareMathSymbol{\au}{\mathalpha}{extraup}{81}
\DeclareMathSymbol{\epfl}{\mathalpha}{extraup}{83}
\DeclareMathSymbol{\wmf}{\mathalpha}{extraup}{84}
\newcommand*\samethanks[1][\value{footnote}]{\footnotemark[#1]}
\author{
Tomás Feith,\thanks{~~Equal contribution, contact: \href{mailto:akhil.arora@cs.au.dk}{akhil.arora@cs.au.dk}.}~$^{\epfl}$
Akhil Arora,\samethanks~\thanks{~~Work done at EPFL.}~$^{\au}$
Martin Gerlach,$^{\wmf}$
Debjit Paul,$^{\epfl}$
Robert West\thanks{~~R. West is a Wikimedia Foundation Research Fellow.}~$^{\epfl}$ \\
$^{\epfl}$EPFL~~~
$^{\au}$Aarhus University~~~
$^{\wmf}$Wikimedia Foundation \\
\texttt{tsfeith@gmail.com}, 
\texttt{akhil.arora@cs.au.dk}, 
\texttt{mgerlach@wikimedia.org}, \\
\texttt{\{debjit.paul, robert.west\}@epfl.ch}
}

\begin{document}
    \maketitle
\begin{abstract}
Links are a fundamental part of information networks, turning isolated pieces of knowledge into a network of information richer than the sum of its parts. However, adding a new link to the network is not trivial: it requires not only the identification of a suitable pair of source and target entities but also the understanding of the content of the source to locate a suitable position for the link in the text. The latter problem has not been addressed effectively, particularly in the absence of text spans in the source that could serve as anchors to insert a link to the target entity. To bridge this gap, we introduce and operationalize the task of \emph{entity insertion} in information networks. Focusing on the case of Wikipedia, we empirically show that this problem is, both, relevant and challenging for editors. We compile a benchmark dataset in 105 languages and develop a framework for entity insertion called \loki (Localized Entity Insertion) and its multilingual variant \xloki. We show that \xloki outperforms all baseline models (including state-of-the-art prompt-based ranking with LLMs such as GPT-4) and that it can be applied in a zero-shot manner on languages not seen during training with minimal performance drop. These findings are important for applying entity insertion models in practice, \eg, to support editors in adding links across the more than 300 language versions of Wikipedia.
\end{abstract}

\section{Introduction}
\begin{figure}[t]
     \centering
     \includegraphics[width=0.9\linewidth]{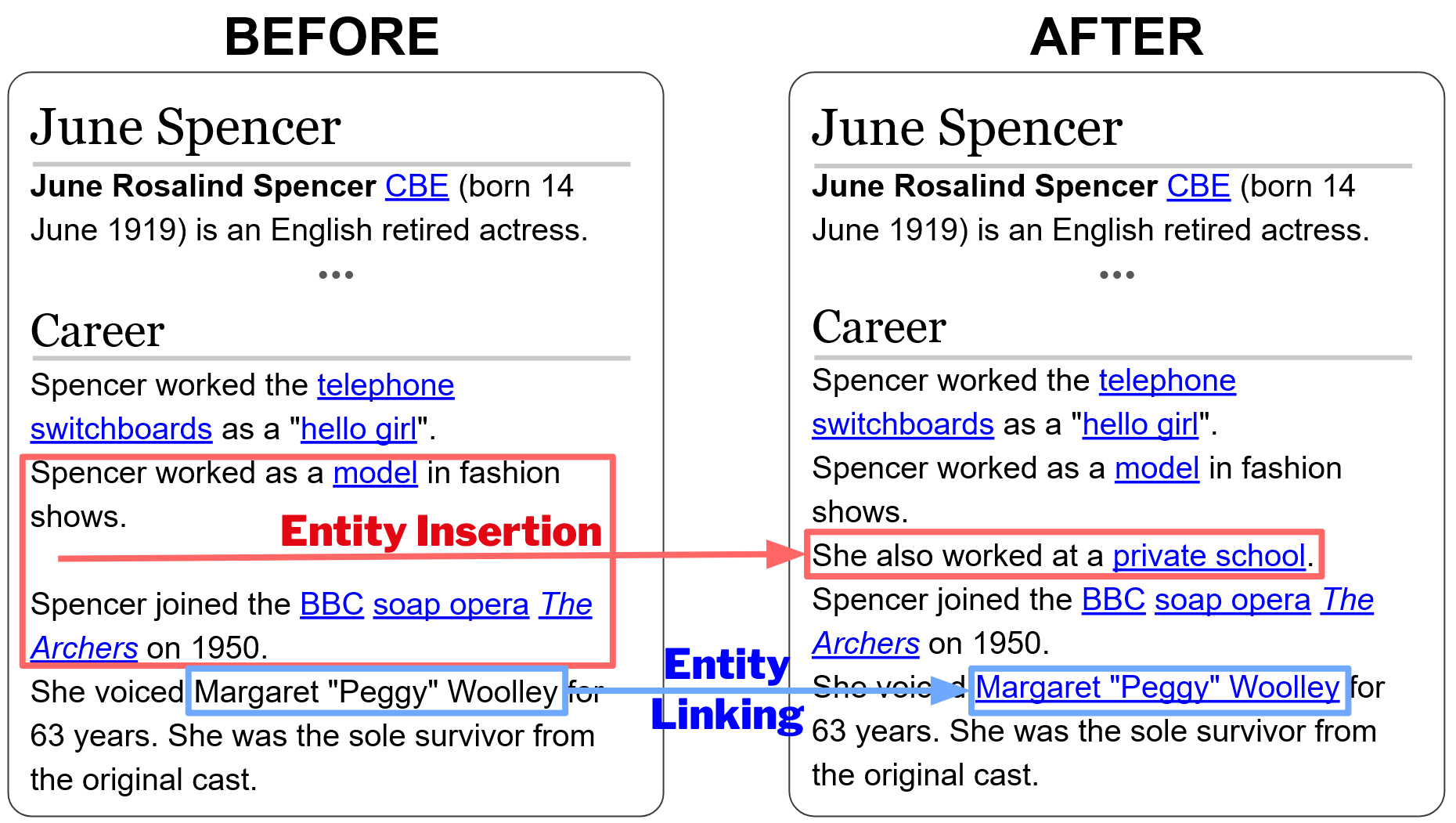}
     \figcaption{\emph{\textcolor{blue}{Entity linking}}: insert a link to the entity \textcolor{black}{\underline{Margaret ``Peggy'' Woolley}} by identifying a suitable mention from the existent text in the version \textbf{before} insertion, \vs \emph{\textcolor{red}{Entity insertion}}: no mention existent yet, identify the most suitable span \fcolorbox{red}{white}{\textcolor{white}{ab}} in the version \textbf{before} to insert the entity \textcolor{black}{\underline{Private school}}.}
     \label{fig:motivation}
     \moveup
\end{figure}

From digital encyclopedias and blogs to knowledge graphs, knowledge on the Web is organized as a network of interlinked entities and their descriptions. 
However, online knowledge is not static: new webpages are created, and existing pages are updated almost every day. 
While there exists substantial support for content creation (\eg via translation~\citet{wulczyn2016growing} or generative AI tools~~\citet{shao2024assisting}), adding new knowledge not only requires creating content but also integrating it into the existing knowledge structure. The latter usually leaves editors with the time-consuming task of reading lengthy webpages to identify a relevant text span for inserting an entity that is not yet mentioned on the page. Thus, to support editors in effectively integrating entities in multilingual linked corpora on the Web, we introduce the task of \emph{entity insertion}.

\xhdr{Entity insertion} We consider Wikipedia as the primary use case and focus on the task of adding links. 
Specifically, given a source and target entity, the goal of \emph{entity insertion} is to identify the most suitable text span in the article describing the source entity for inserting a link to the target entity. Fig.~\ref{fig:motivation} portrays a real example of the entity insertion task with the eventual goal of adding a link from the source entity \texttt{June Spencer}, a former English actress, to the target entity \texttt{Private school}. 
Most importantly, entity insertion is a different and much more challenging task when compared to \emph{entity linking}, as no existent text span in the version of the source article (\texttt{June Spencer}) at edit time could be used to link to the target entity (\texttt{Private school}). Rather, a new text span--\emph{``She also worked at a private school.''}--was added along with the to-be-inserted target entity.

\begin{figure}[t]
    \includegraphics[width=0.485\linewidth]{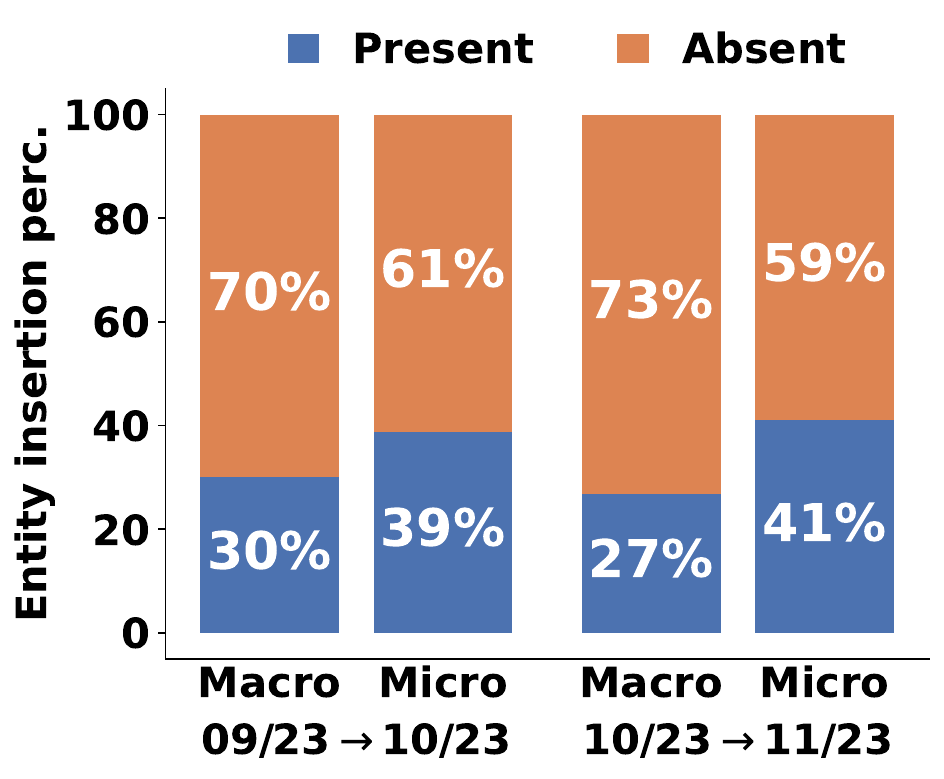}
    \hfill
    \includegraphics[width=0.485\linewidth]{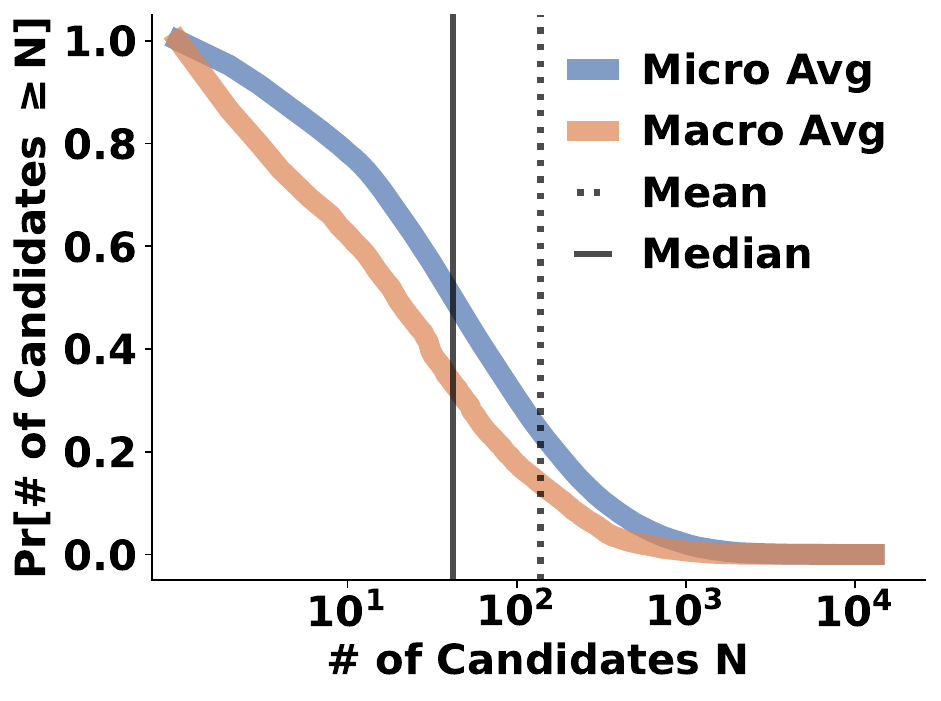}
    \figcaption{Challenges of entity insertion. (Left) Micro (weighted by the number of data points in a language) and macro (equal weight to each language) aggregates of insertion types over the 105 languages considered in this study. (Right) Complementary cumulative distribution function (CCDF) of the number of candidate sentences ($N$) in a Wikipedia article (log x-axis).}
    \label{fig:challenges}
    \moveups
\end{figure}

\xhdr{Challenges} Entity insertion is not only an interesting and challenging language understanding task, but it is also the most common scenario faced by editors when adding links in practice. 
In fact, we find that for \emph{60-70\%} of all the links added to Wikipedia, \emph{none of the existing text} is suitable to insert the corresponding entities, and new text needs to be added along with the entity by the editor (Fig.~\ref{fig:challenges}). 
Fig.~\ref{fig:challenges} also shows that entity insertion is associated with a \emph{high cognitive load}, as the task requires, on average, an editor to select the most suitable sentence from a pool of $\sim$100 candidate sentences.

Therefore, it is vital to operationalize and develop new methods for entity insertion in order to support editors in adding links to Wikipedia and other information networks. 
To this end, we make the following key contributions in this paper.

\xhdr{Contributions}
We introduce the \emph{novel task of entity insertion} 
(\S~\ref{sec:entity_insertion}). 
We release a \emph{large dataset in 105 languages} of links from Wikipedia articles to enable further research into entity insertion (\S~\ref{sec:data}). 
We introduce \loki, a framework for entity insertion, and its multilingual variant \xloki (\S~\ref{sec:method}). 
We show the benefit of multilingual knowledge in downstream performance and highlight the \emph{zero-shot} capabilities of \xloki (\S~\ref{sec:results}).

\section{Related work}
In this section, we review works that overlap closely with our study (\cf Appx.~\ref{app.additional_related_work} for details).

\xhdr{Entity linking}
\label{sec:background_entity_linking}
Previous work has framed entity insertion as an entity linking problem~\citep{entity_linking_1, entity_linking_3, entity_linking_4, entity_linking_5, entity_linking_6,entity_linking_7}, where the goal is to assign a unique identity to entities mentioned in the text. The task of entity linking is composed of two sub-tasks: Named Entity Recognition (NER) and Named Entity Disambiguation (NED). Most research \citep{me_tp_1, me_tp_2, me_tp_3} into entity linking solves first the NER problem, in which the task is to find candidate mentions for named entities in the source article. 
However, there is recent work \citep{tp_me_1} exploring the problem in reverse order, first solving NED by finding target entities related to the source article and then NER searching only for mentions for the found targets. 

When the mention is present, the task of entity insertion is similar to NER~\cite{tp_me_1}, as both tasks can be solved by searching for mentions in the text. 
However, entity insertion is a more general task as it allows for the mention of the target entity to not yet be present in the text. In this case, the goal is to exploit the context information to find the text span most related to the target entity. NER modules \citep{ner_3, ner_1} are designed to search for the most related mentions, and thus, they are not applicable in scenarios where the mentions are not yet available.

\xhdr{Entity tagging} 
\citet{entity_tagging} introduced this task as a relaxed form of entity linking. 
An entity tagging model is only tasked with determining the entities present in the text and does not need to find the exact mentions of the entities. 
However, even though the model is not tasked with extracting an entity's mention, the task of entity tagging still assumes that the text contains some mention of the entity, which distinguishes it from entity insertion.

\xhdr{Link the Wiki}
\citet{link_the_wiki} ran a track at INEX 2008 with two tasks: file-to-file link discovery and mention-to-BEP (best entry point) link discovery. 
File-to-file link discovery is a document-level task that can be framed as a link prediction task in networks, where the Wikipedia articles act as nodes and the links act as edges. 
The mention-to-BEP task is an entity linking task with anchor prediction, where the two-part goal is to find mentions in the source article pointing to other articles, and finding the best point of entry (the anchor) in the target file. 
This task has more recently resurfaced as an anchor prediction task \citep{anchor_prediction}. 

\begin{figure*}[!htb]
    \centering
    \includegraphics[width=0.93\textwidth]{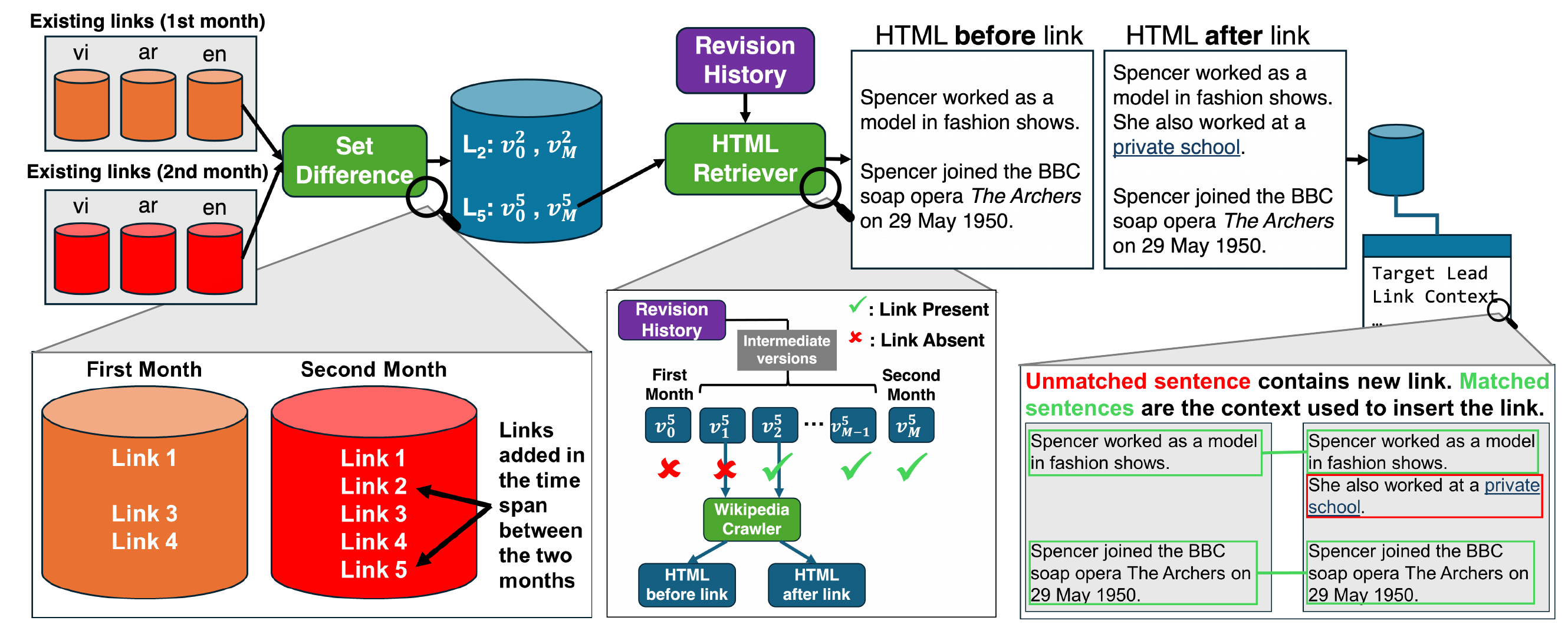}
    \figcaption{Data processing pipeline. Obtain added links $L$ by taking a set difference of the links existent in consecutive months. For each added link $L_i$, scan all $M$ versions in the full revision history $v_0^i$ to $v_M^i$ to identify the article version in which the link was added and compute the difference between the before and after versions to extract the exact entity insertion scenario.}
    \label{fig:data_pipeline}
\end{figure*}

\xhdr{Passage ranking} 
Transformer-based models have revolutionized passage ranking by enhancing semantic understanding beyond traditional lexical methods like BM25 \cite{bm25}. BERT demonstrated early success by leveraging contextualized embeddings for re-ranking \cite{bert_passage_rank}, leading to innovations like ColBERT \cite{colbert}, which uses a dual-encoder architecture for more efficient retrieval. Recent models such as T5 \cite{t5} and ELECTRA \cite{electra} further refine ranking by employing advanced pre-training techniques. Building on top of this work, \cite{kgpr, kg_rerank} employ knowledge graphs to exploit background information to better rank passages. However, such graph-based approaches are not suited for large-scale, highly dynamic graphs (such as Wikipedia), as the cost of recomputing all the embeddings associated with the graph is too high. Finally, while large language models have been shown to be the state of the art for passage ranking~\cite{qin2024large}, despite their performance they are impractical at the Web-scale owing to exorbitantly high computational costs.

\xhdr{Key differences}
Entity insertion is fundamentally different from all the aforementioned tasks and possesses novel downstream applications. 
First, entity insertion does not assume that a mention to the target entity is present in the text at inference time. 
Second, the optimization objective of entity insertion, which involves identifying the text span most related to a target entity, could be seen as the dual of tasks such as NED and entity tagging, which aim instead to find the most relevant target entity for a given text span. 
Finally, entity insertion aims to find the best text span in the source article to insert the target entity. 
In contrast, anchor prediction performs the reverse task by trying to find the best text span for grounding the source entity in the target article. 
Moreover, anchor prediction is an unnatural task as humans find the vast majority of links to be unanchorable~\citep{anchor_prediction}.

\section{Task formulation}
\label{sec:entity_insertion}
Let $E_{\text{src}}$ be a source entity and $E_{\text{tgt}}$ be a target entity. Let $X_{src}$ be the textual content of the article corresponding to $E_{\text{src}}$. The text can be partitioned into a set of (potentially overlapping) text spans, $\mathcal{X}_{src}=\left\{x_1,...,x_M\right\}$, where $M$ is the number of text spans in the article. 
Entity insertion is the task of selecting the most relevant span $x^*$ to insert the target entity $E_{\text{tgt}}$. Formally, 
\moveup
\begin{equation}
\label{eq:entity_insertion}
x^*=\arg \max_{x\in\mathcal{X}_{src}}\mathcal{R}(x, E_{\text{tgt}})
\moveup
\end{equation}
where $\mathcal{R}$ is an arbitrary relevance function quantifying the relevance of $E_{\text{tgt}}$ to each text span $x\in\mathcal{X}_{src}$. 
We frame entity insertion as a \emph{ranking task}, where the goal is to rank all the candidate text spans $\mathcal{X}_{src}$ based on their relevance to the target entity.

\section{Data}
\label{sec:data}
We constructed a new multilingual dataset for studying entity insertion in Wikipedia. 
The dataset consists of links extracted from all Wikipedia articles, each link's surrounding context, and additional article-level meta-data (such as titles, Wikidata QIDs, and lead paragraphs). 
Overall, the dataset contains 958M links from 49M articles in 105 languages. 
The largest language is English (en), with 166.7M links from 6.7M articles, and the smallest language is Xhosa (xh), with 2.8K links from 1.6K articles 
(\cf Appendix~\ref{app.data} for details).

Fig.~\ref{fig:data_pipeline} provides an overview of our data processing pipeline. The data processing was done in two steps.  
We first extracted all the links from the 2023-10-01 snapshot. 
Next, we found all the links added in the time between 2023-10-01 and 2023-11-01. 

\begin{figure*}[t]
    \moveup
     \centering
     \includegraphics[width=0.88\textwidth]{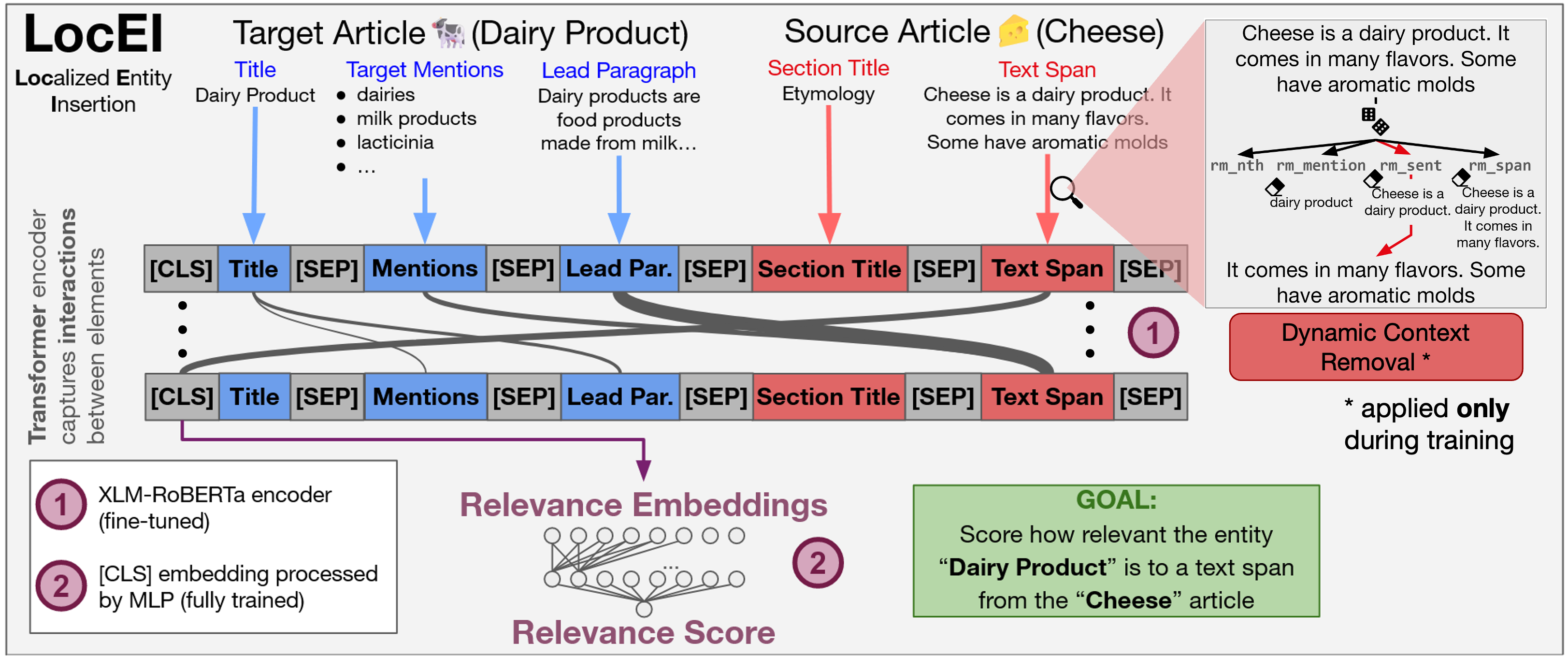}
     \figcaption{Architectural overview of \loki. The target entity $E_{tgt}$ and each candidate text span $x \in \mathcal{X}_{src}$ of the source entity $E_{src}$ are concatenated together and encoded jointly using a transformer encoder. The relevance scores of candidate text spans are computed using an MLP trained via a list-wise ranking objective.}
     \label{fig:model_architecture}
\end{figure*}

\xhdr{Existing links}
\label{sec:existing_links}
We extract the content of all articles from their HTML version using the corresponding snapshot of the Enterprise HTML dumps~\cite{htmldumps}.
We removed articles without a lead paragraph and a Wikidata QID.
For each article, we consider all internal links in the main article body (ignoring figures, tables, notes, and captions) together with their surrounding context. 
We removed all the links where either the source or the target article was one of the removed articles and we dropped all the self-links.

\xhdr{Added links}
We extract the set of added links by comparing existing links in snapshots from consecutive months. 
We apply the same procedure as above to each snapshot, respectively, and take the difference of the two sets to identify the links that exist in the second month but not in the first.

To identify the article version in which the link was added, we go through the articles' full \emph{revision history} available in the Wikimedia XML dumps~\cite{wikipedia_xmldumps}. 
Next, we identify the two versions of an article before and after the link addition and download the corresponding HTML. 
Comparing the two HTML versions, we extract the content modifications made by the editor when adding the link, and categorize them into \emph{five entity insertion scenarios}. 
(1) \texttt{text\_present}: the link was added by hyperlinking an existing mention; (2) \texttt{missing\_mention}: the link was added by adding the mention for a new entity (and potentially some additional content) into an existent sentence; (3) \texttt{missing\_sentence}: the link was added by writing a new sentence to complement the already existing text and hyperlinking part of the sentence at the same time; (4) \texttt{missing\_span}: an extension of the previous category, where the editors added a span of multiple sentences; and (5) \texttt{missing\_section}: the link was added in a section that did not exist in the previous version of the article. 
We provide examples (Table~\ref{tab:link_insertion_examples}) and frequency of occurrence (Fig.~\ref{fig:full_link_distributions}) of these cases in Appendix~\ref{app.link_insertion_examples}. 

\xhdr{Data release} The dataset is made publicly available on Zenodo under an open license (CC-BY-SA 4.0) at \url{https://zenodo.org/records/13888211}. 

\section{Entity insertion with \loki}
\label{sec:method}
Fig.~\ref{fig:model_architecture} presents an overview of \loki. Our model (\S~\ref{sec:architecture}) is composed of a transformer-based encoder that jointly encodes the target entity as well as the candidate spans in the source entity, and a multilayer perceptrion (MLP) trained via a list-wise objective capable of ranking candidates based on their relevance to the target. We introduce a novel data augmentation strategy that closely mimics real-world entity insertion scenarios (\S~\ref{subsec:augmentation}), a knowledge injection module to incorporate external knowledge about the entities (\S~\ref{subsec:knowledge_injection}), and the multilingual variant \xloki (\S~\ref{subsec:xloki}). 

\subsection{Model}
\label{sec:architecture}
\xhdr{Architecture} We use a transformer-based encoder $\Gamma$ to jointly encode each candidate text span $x \in \mathcal{X}_{src}$ and the target entity $E_{tgt}$ into a sequence of vectors. To reduce this sequence into a single vector, we use the embedding of the \texttt{[CLS]} token~\cite{bert}, which measures how related the candidate $x$ is to the entity $E_{tgt}$. An MLP $\Lambda$ produces a scalar relevance score between the candidate $x$ and the entity $E_{tgt}$ using the relevance embedding produced by the encoder $\Gamma$ defined as  
\moveup
\begin{equation}
\moveups
\label{eq:vector_relevance}
    R=\Gamma\left(\phi; \theta_{\Gamma}\right)
\moveup
\end{equation}
\begin{equation}
\moveups
\label{eq:scalar_relevance}
    r=\Lambda\left(R_{\texttt{[CLS]}}; \theta_{\Lambda}\right)  
\end{equation}
where $\Gamma$ is an encoder and $\Lambda$ is an MLP with $\theta_{\Gamma}$ and $\theta_{\Lambda}$ as the learnable parameter spaces, respectively, $\phi$ is obtained by concatenating the input representations of $E_{tgt}$ and $x$, $R$ is the sequence of d-dimensional contextualized embeddings produced by $\Gamma$, $R_{\texttt{[CLS]}}$ is the $d$-dimensional relevance embedding, and $r$ is the relevance scalar produced by $\Lambda$ to rank the candidates.

\xhdr{Entity and candidate span modeling}
We represent the target entity $E_{tgt}$ via two textual features, the title $T_{tgt}$ and the lead text $L_{tgt}$ which is a short paragraph present in most Wikipedia articles. Each candidate span $x$ is represented via the text $t$ contained in $x$. These textual features are concatenated together into a single textual input $\phi$ as, 
\moveup
\begin{equation}
\moveups
    \phi=\mathcal{T}(\texttt{[CLS]}T_{tgt}\texttt{[SEP]}L_{tgt}\texttt{[SEP]}t\texttt{[SEP]})
\label{eq:base_input}
\moveup
\end{equation}
where $\mathcal{T}(\cdot)$ is the tokenizer operator that produces a sequence of $T$ tokens.

\xhdr{Optimization}
Given that entity insertion is a ranking task, we use an objective function that introduces the notion of ranking into the model. Specifically, we train the relevance scoring module using a cross-entropy loss over a \emph{list} of candidates. Given a target entity $E_{tgt}$, a list of $N$ candidate text spans $\mathcal{X}_N=[x_1,\dots,x_N]$, and $i'$ as the index of the correct candidate, we use the following list-wise objective, 
\moveup
\begin{equation*}
\moveups
    \max_{\theta}\frac{\exp\left(\operatorname{score}\left(x_{i'}, E_{tgt}; \theta\right)\right)}{\sum_{i=1}^N\exp\left(\operatorname{score}\left(x_i, E_{tgt}; \theta\right)\right)}
\end{equation*}
where $\operatorname{score}$ is an operator chaining the operations from Equations~\ref{eq:vector_relevance} and \ref{eq:scalar_relevance}.

\xhdr{Inference}
The document $X_{src}$ in which to insert the entity $E_{tgt}$ may contain a number $D$ of potentially overlapping text spans $\mathcal{X}_{src}=[x_1,...,x_D]$. At inference time, the procedure described above is applied to all the $D$ candidate text spans, and a relevance score is obtained for each candidate.

\subsection{Two-stage training pipeline}
\label{subsec:augmentation}
We extract two types of links for studying entity insertion: existing and added links (\S~\ref{sec:data}). While added links reflect the entity insertion scenarios observed in the real world, we found that the number of added links is low for most languages (\cf Table~\ref{tab:data_statistics_train_test} in the Appendix), thereby not being sufficient for training our model. To circumvent this challenge, we develop a \textit{two-stage training pipeline} that uses both existing and added links.

\xhdr{Dynamic context removal}
A key challenge with existing links is that they only reflect the \texttt{text\_present} category of entity insertion, as the mention of the target entity is always present in the article containing the link. We mitigate this challenge by introducing a novel \emph{data augmentation} strategy to simulate all other real-world entity insertion scenarios that are missing in the existing links. \emph{Dynamic context removal} modifies the context associated with each existing link during training to simulate editors' edits of adding links under different scenarios of entity insertion discussed in~\S\ref{sec:data}. Specifically, to simulate the \texttt{missing\_mention}, \texttt{missing\_sentence}, and \texttt{missing\_span} scenarios, we randomly remove a word (\texttt{rm\_mention}), a sentence (\texttt{rm\_sent}), or a span of sentences (\texttt{rm\_span}), respectively. 
Table~\ref{tab:context_masking_description} summarizes the strategies (\cf Table~\ref{tab:context_mask} in Appx.~\ref{app.dynamic_context_removal} for details with examples). 

Note that dynamic context removal may generate structural and linguistic patterns that would not occur in the text written by human editors. For example, applying the \texttt{rm\_mention} strategy on the sentence ``\emph{Laika was a \underline{Soviet space dog} who was one of the first animals in space to orbit the Earth.}'', would produce the sentence ``\emph{Laika was a who was one of the first animals in space to orbit the Earth.}''. Such a sentence is unlikely to be found in natural text articles, and thus, there is a distribution shift from the augmented training data to the test data.

\xhdr{Expansion} To reduce the impact of this shift, we introduce a second stage of training where we use the added links containing real-world entity insertion scenarios. Note that unlike the first stage, which uses existing links, the second stage does not require dynamic context removal, as we have access to the real contexts used by editors covering all the entity insertion scenarios. 

\begin{table}
    \centering
    \vspace{3mm}
    \tabcaption{Dynamic context removal strategies.}
    \vspace{1mm}
    {\renewcommand{\arraystretch}{1.2}
    \resizebox{0.4\textwidth}{!}{%
    \begin{tabular}{c|c}
        \toprule
        Strategy & Text Removed \\\hline
        \texttt{rm\_nth} & None \\
        \texttt{rm\_mention} & Mention \\
        \texttt{rm\_sent} & Sentence containing mention \\
        \texttt{rm\_span} & Span of sentences containing mention \\\hline
    \end{tabular}
    }
    }
    \label{tab:context_masking_description}
    \moveup
    \moveup
    \moveup
\end{table}

\subsection{Knowledge injection}
\label{subsec:knowledge_injection}
While the representation presented in Eq.~\ref{eq:base_input} (\S~\ref{sec:architecture}) already allows \loki to measure the target entity's relevance to the candidate text span, we inject external knowledge about the target entity and knowledge about the structural organization of the source article to produce better relevance embeddings.

Since section titles provide additional `local' knowledge in the form of a summarized conceptualization of a candidate span, we first add the title of the section $s$ in which a span appears to its input representation. 
Next, we add the list of mentions $M_{tgt}$ previously associated with the target entity. This list provides a strong signal of how the entity is typically referenced in the text, thereby facilitating the model to better attend to these mentions when computing the relevance embedding. 
The final input format after knowledge injections is: 
\moveup
\moveups
\begin{equation*}
\moveups
    \resizebox{0.99\linewidth}{!}{%
        $\phi=\mathcal{T}(\texttt{[CLS]}T_{tgt}\space M_{tgt}\texttt{[SEP]}L_{tgt}\texttt{[SEP]}s\texttt{[SEP]}t\texttt{[SEP]})$%
        }
\label{eq:final_input}
\end{equation*}

\subsection{Incorporating multilinguality (\xloki)}
\label{subsec:xloki}
To enable the encoder to better model the relationship between an entity target and candidate text spans, we leverage the patterns existent in multiple languages. 
For this, we train a single model by jointly considering entity insertion examples in multiple languages. 
This enables cross-lingual transfer, empowering, especially, low-resource languages with lesser and lower quality training data.

\begin{table*}[t]
    \centering
    \tabcaption{Entity insertion performance obtained by macro-averaging over 20 Wikipedia language versions used for training the benchmarked methods. \xloki trains a single model jointly on all 20 languages, whereas other methods train a separate model for each language. The categorization of entity insertion types into `Overall', `Missing', and `Present' is discussed in \S~\ref{subsec:setup}. Note that EntQA and GET work only for English (results in Table~\ref{tab:multilingual_results_extra}), whereas PRP-Allpair was only used for zero-shot analysis (Table~\ref{tab:zero_shot}) and English (Table~\ref{tab:multilingual_results_extra}).}
    \vspace{1mm}
    \begin{threeparttable}
    \resizebox{0.8\linewidth}{!}{
        \begin{tabular}{c|c|ccc|ccc}
         \toprule
         & \multirow{2}{*}{Method} & \multicolumn{3}{c|}{Hits@1} & \multicolumn{3}{c}{MRR} \\
         & & Overall & Present & Missing & Overall & Present & Missing \\\hline
        Baseline & Random & 0.107 & 0.115 & 0.103 & 0.243 & 0.259 & 0.236 \\
        Baseline & String Match & 0.459 & 0.708 & 0.270 & 0.557 & 0.774 & 0.395 \\
        Baseline & BM25 & 0.508 & 0.799 & 0.280 & 0.612 & 0.866 & 0.421 \\
        Baseline & Simple fine-tuning & 0.584 & 0.883 & 0.350 & 0.649 & 0.907 & 0.451 \\\hline
        Proposed & \loki & 0.672 & 0.877 & 0.509 & 0.744 & 0.906 & 0.617 \\
        Proposed & \xloki & \textbf{0.726}$^{\dagger}$ & \textbf{0.909}$^{\dagger}$ & \textbf{0.579}$^{\dagger}$ & \textbf{0.789}$^{\dagger}$ & \textbf{0.929}$^{\dagger}$ & \textbf{0.678}$^{\dagger}$ \\\hline
        \end{tabular}
        }
    \begin{tablenotes}
        \small \item[\dag] Indicates statistical significance ($p<0.05$) between the best and the second-best scores.
    \end{tablenotes}
    \end{threeparttable}
    \label{tab:multilingual_results}
\end{table*}

\begin{table*}[t]
    \centering
    \vspace{1mm}
    \tabcaption{Entity insertion performance obtained for English.}
    \vspace{1mm}
    \begin{threeparttable}
    \resizebox{0.8\linewidth}{!}{
        \begin{tabular}{c|c|ccc|ccc}
        \toprule
         & \multirow{2}{*}{Method} & \multicolumn{3}{c|}{Hits@1} & \multicolumn{3}{c}{MRR} \\
         & & Overall & Present & Missing & Overall & Present & Missing \\\hline
        Baseline & Random & 0.079 & 0.110 & 0.067 & 0.202 & 0.240 & 0.187 \\
        Baseline & String Match & 0.391 & 0.732 & 0.264 & 0.489 & 0.796 & 0.374 \\
        Baseline & BM25 & 0.439 & 0.838 & 0.290 & 0.538 & 0.894 & 0.404 \\
        Baseline & EntQA$_{\text{RET}}$ & 0.099 & 0.136 & 0.085 & 0.234 & 0.278 & 0.217 \\
        Baseline & GET & 0.391 & 0.827 & 0.228 & 0.469 & 0.851 & 0.326 \\
        Baseline & PRP-Allpair (GPT-3.5)~\cite{qin2024large} * & 0.160 & 0.375 & 0.092 & 0.322 & 0.536 & 0.255 \\
        Baseline & PRP-Allpair (GPT-4)~\cite{qin2024large} * & 0.370 & 0.833 & 0.224 & 0.499 & 0.877 & 0.380 \\
        Baseline & Simple fine-tuning & 0.443 & 0.860 & 0.287 & 0.522 & 0.888 & 0.385 \\\hline
        Proposed & \loki & \textbf{0.677}$^{\dagger}$ & \textbf{0.879} & \textbf{0.602}$^{\dagger}$ & \textbf{0.741}$^{\dagger}$ & \textbf{0.902} & \textbf{0.681}$^{\dagger}$ \\\hline
        \end{tabular}
        }
    \begin{tablenotes}
        \small \item[\dag] Indicates statistical significance ($p<0.05$) between the best and the second-best scores.
        \item[*] Evaluation on a sample of 100 test instances.
    \end{tablenotes}
    \end{threeparttable}
    \label{tab:multilingual_results_extra}
    \moveup
    \moveup
    \moveup
\end{table*}

\section{Experiments}
\label{sec:results}
All the resources required to reproduce the experiments in this paper are available at \url{https://github.com/epfl-dlab/multilingual-entity-insertion}.

\subsection{Data}
\label{exp:data}
We study entity insertion in 105 language versions of Wikipedia. We use a judicious mix (based on size, script, geographic coverage, \etc) of 20 languages for training the benchmarked methods, however, for evaluation, we consider all 105 languages. For dataset statistics, \cf Tables~\ref{tab:data_statistics} and~\ref{tab:data_statistics_train_test} of Appx.~\ref{app.data}.

\xhdr{Training set} 
We train \loki and \xloki using a two-stage training pipeline (\S~\ref{subsec:augmentation}). While the data for the first stage is based on the existing links extracted from the 2023-10-01 snapshot, the second stage data is built using the links added between the 2023-09-01 and 2023-10-01 snapshots. 

\xhdr{Negative candidates}
\label{sec:neg_cands}
During training, we extract $N$ negative candidates for each positive candidate. Negative candidates are text spans in the source $X_{src}$ where the target entity $E_{tgt}$ was not inserted. Whenever possible, we select $N$ negative candidates (``hard negatives'') from the same source article as the positive candidate. However, when articles are too small to be able to select $N$ negatives, we sample the remaining negative candidates randomly from other articles (``easy negatives''). Details pertaining to the implementation of negative candidate extraction are provided in Appendix~\ref{app.negative_sampling}.

\xhdr{Test set} For evaluation, we use the links added between the 2023-10-01 and 2023-11-01 snapshots. This ensures no overlap between the training and test sets and is therefore advantageous in mitigating data leakages. 
Unlike training, we use all the $D$ available candidates in an article for evaluation.

\subsection{Baselines}
\label{subsec:baselines}

\noindent $\bullet$ \textbf{Random:} ranks candidates uniformly at random.

\noindent $\bullet$ \textbf{String Match:} searches for previously used mentions in the candidate text spans.

\noindent $\bullet$ \textbf{BM25~\cite{bm25}:} applies the Okapi-BM25 implementation \cite{bm25_implementation} on keywords extracted from the target lead paragraph and the candidate text spans.

\noindent $\bullet$ \textbf{EntQA~\cite{tp_me_1} (English only):} for independently encoding the candidate text spans and target entity. We then use the retriever model of EntQA to rank text spans based on the cosine similarity between the embeddings.

\noindent $\bullet$ \textbf{GET~\cite{entity_tagging} (English only):} use the generative ability of GET to generate the target entity name for each candidate text span. We then rank the text spans based on their likelihood of generating the target entity.

\noindent $\bullet$ \textbf{PRP-Allpair~\cite{qin2024large} (Zero-shot only):} to assess the relevance of candidate text spans to the target entity in a pairwise manner using GPT-3.5 and GPT-4, and then uncover the ranking from all pairwise comparisons.

\subsection{Setup}
\label{subsec:setup}
\xhdr{Model} We present results for \loki and \xloki by fine-tuning the pre-trained \texttt{xlm-roberta-base} model~\cite{xlm_roberta} as the encoder. The MLP is a 2-layer network with ReLU activations. We also explored different model sizes (\eg Large and XL) and other pre-trained models (BERT and T5): results in Appendix~\ref{app:additional_experiments}.

\begin{table*}[t]
    \centering
    \tabcaption{Entity insertion performance in the zero-shot setting: results obtained by macro-averaging over 9 Wikipedia language versions that were not used for fine-tuning \xlokismall. \xlokilarge was trained jointly on all 20 languages, whereas \loki trains a separate model for each language. The categorization of entity insertion types into `Overall', `Missing', and `Present' is discussed in \S~\ref{subsec:setup}.}
    \vspace{1mm}
    \label{tab:zero_shot}
    \begin{threeparttable}
    \resizebox{0.8\linewidth}{!}{
        \begin{tabular}{c|c|ccc|ccc}
        \toprule
         & \multirow{2}{*}{Method} & \multicolumn{3}{|c|}{Hits@1} & \multicolumn{3}{|c}{MRR} \\
         & & Overall & Present & Missing & Overall & Present & Missing \\\toprule
        Fine-tuned & \loki & 0.647 & 0.873 & 0.486 & 0.718 & 0.902 & 0.588 \\
        Fine-Tuned & \xlokilarge & \textbf{0.709}$^{\dagger}$ & \textbf{0.901} & \textbf{0.570}$^{\dagger}$ & \textbf{0.772}$^{\dagger}$ & \textbf{0.923} & \textbf{0.662}$^{\dagger}$ \\   
        \midrule
        Zero-shot & PRP-Allpair (GPT-3.5)~\cite{qin2024large} * & 0.289 & 0.423 & 0.210 & 0.433 & 0.563 & 0.353 \\
        Zero-shot & PRP-Allpair (GPT-4)~\cite{qin2024large} * & 0.571 & 0.859 & 0.344 & 0.656 & 0.897 & 0.468 \\
        Zero-Shot & \xlokismall & \bf 0.690$^{\dagger}$ & \bf 0.887 & \bf 0.541$^{\dagger}$ & \bf 0.755$^{\dagger}$ & \bf 0.913 & \bf 0.636$^{\dagger}$ \\
        \bottomrule
        \end{tabular}
    }
    \begin{tablenotes}
        \small \item[\dag] Indicates statistical significance ($p<0.05$) from fine-tuned \loki. 
        \item[*] Evaluation on a sample of 100 test instances.
    \end{tablenotes}
    \end{threeparttable}
    \moveup
    \moveup
    \moveup
\end{table*}

\begin{table}[b]
    \centering
    \moveup
    \tabcaption{Entity insertion performance in the full zero-shot setting: results obtained by macro-averaging over 85 held-out Wikipedia language versions that were not used for fine-tuning the benchmarked methods.}
    \label{tab:language_scaling}
    \begin{threeparttable}
        \resizebox{0.99\linewidth}{!}{
            \begin{tabular}{l|ccc|ccc}
            \toprule
            \multirow{2}{*}{Method} & \multicolumn{3}{c|}{Hits@1} & \multicolumn{3}{c}{MRR} \\
             & Overall & Present & Missing & Overall & Present & Missing \\\hline
            Random & 0.148 & 0.132 & 0.148 & 0.288 & 0.287 & 0.281 \\
            String Match
             & 0.442 & 0.717 & 0.273 & 0.549 & 0.786 & 0.406 \\
            BM25 & 0.456 & 0.733 & 0.294 & 0.580 & 0.823 & 0.435 \\\hline
            \xlokismall & 0.683 & 0.853 & 0.585 & 0.754 & 0.886 & 0.676 \\
            \xlokilarge & \textbf{0.706}$^{\dagger}$ & \textbf{0.873}$^{\dagger}$ & \textbf{0.602} & \textbf{0.769} & \textbf{0.901} & \textbf{0.685} \\\hline
            \end{tabular}
        }
    \begin{tablenotes}
        \small \item[\dag] Indicates statistical significance ($p<0.05$) between \\ the best and the second-best scores.
    \end{tablenotes}
    \end{threeparttable}
    \moveup
\end{table}

\xhdr{Evaluation metrics} We use (1) Hits@1, and (2) mean reciprocal rank (MRR) to evaluate the quality of the benchmarked methods. For each language, we compute the micro aggregates of the metrics over all added links in the test set. 
Moreover, we present results grouped into three categories: (1) Overall: considering the entire test set, (2) Present: considering links corresponding to the \texttt{text\_present} entity insertion scenario, and (3) Missing: considering links corresponding to all the other scenarios, namely, \texttt{missing\_mention}, \texttt{missing\_sentence}, and \texttt{missing\_span}.

Additional details about the experimental setup and hyperparameter tuning (impact of pre-trained models, model sizes, training stages, pointwise \vs ranking loss, \etc) are present in Appendix~\ref{app:additional_experiments}.

\subsection{Main results}
\label{sec:multilingual_results}

We evaluate three variants of our entity insertion model: i) \textit{simple fine-tuning}: a family of monolingual models fine-tuned in each language without the extensions (data augmentation, knowledge injection, two-stage training) introduced in \loki; ii) \loki: a family of monolingual models fine-tuned using the full \loki framework; and iii) \xloki, a single multilingual model fine-tuned jointly on all the languages using the full \loki framework. 
Table~\ref{tab:multilingual_results} shows the models' performance metrics (Hits@1 and MRR) aggregated (macro-average) over the 20 considered languages. 

\xhdr{Overall performance}
We see that \xloki achieves the best overall quality and statistically significantly outperforms all other models for all cases considered. The key highlights are as follows: (1) \emph{BM25}, a hard-to-beat baseline for ranking tasks, is around $20$ percentage points inferior to \xloki, (2) \emph{simple fine-tuning}, a baseline that we introduce in this work, substantially outperforms all the other considered methods, but is inferior to \loki and \xloki by being about $10$ and $15$ percentage points worse, respectively, and (3) \xloki consistently yields better scores than the language-specific \loki models, demonstrating that the multilingual model is capable of transferring knowledge across languages to improve overall performance. In fact, by looking at the performance for the individual languages in Figs.~\ref{fig:full_hits_at_1} and \ref{fig:full_mrr} (Appx.~\ref{app.main_result_by_language}), we see that the improvement from \xloki over \loki is larger in low-resource languages (languages with less training data) such as Afrikaans (af), Welsh (cy), Uzbek (uz).

\xhdr{Performance on `Missing' and `Present' categories}
The key finding is that the baselines lack robustness to the variation in entity insertion types, which is substantiated by the huge disparity of entity insertion performance (around $50$ percentage points) of all the baselines in the `Present' and `Missing' categories. This result further highlights the key limitation of the baselines: they cannot address the challenging scenarios of entity insertion. The key reason behind this disparity is that all the existing baselines rely on the existence of a suitable text span to insert a link to the target entity. On the contrary, both \loki and \xloki effectively utilize the signals manifested in the context due to the introduced extensions (\eg data augmentation) and are therefore robust to different entity insertion scenarios. Consequently, we observe that both \loki and \xloki obtain substantial improvements over all the baseline models in the \texttt{missing} category.

\xhdr{Performance on English} \tomas{Table~\ref{tab:multilingual_results_extra} shows that even in English (a high-resource language), \xloki outperforms all baselines. Once again, this gap is pronounced in the \texttt{missing} case, further highlighting the difficulty and novelty of the task.}

\subsubsection{Zero-shot \vs\ Fine-tuned}
\label{sec:zero_shot}

We further study the performance of \xloki in the zero-shot scenario, \ie, evaluating the model in languages that were not explicitly contained in the data for fine-tuning.
This is relevant to assess the potential to support languages for which there is little or no training data available. 
We consider \xlokismall, a variant of the multilingual \xloki which is trained on only 11 out of the 20 languages (\cf Table~\ref{tab:zero_shot_models} in Appx.~\ref{app.zeroshot_result_by_language} for details). 
We then evaluate the zero-shot performance of \xlokismall in the remaining 9 languages not considered for training. 
For comparison, we also show the non-zero shot performance of the models considered in the previous subsection: i) \loki, the family of monolingual models fine-tuned in each language; and ii) \xlokilarge, the single multilingual model trained on all 20 languages. 
The main result, shown in (Table~\ref{tab:zero_shot}), is that \xlokismall retains over 95\% performance in the zero-shot scenario in comparison to the results of the best model, \xlokilarge, which was fine-tuned on these languages. 
Nevertheless, \xlokismall still outperforms the language-specific \loki models fine-tuned on each language individually. 
We expand the robustness of these results by considering two additional scenarios.

\begin{table*}[t]
    \centering
    \tabcaption{Analyzing the impact of the extensions introduced in the \loki framework on the entity insertion performance for only English and the macro-average over 20 Wikipedia language versions. The categorization of entity insertion types into `Overall', `Missing', and `Present' is discussed in \S~\ref{subsec:setup}.}
    \vspace{1mm}
    \label{tab:novelties_analysis}
    \begin{threeparttable}
        \resizebox{0.95\linewidth}{!}{
            \begin{tabular}{l|ccc|ccc|ccc|ccc}
            \toprule
             & \multicolumn{6}{c|}{English} & \multicolumn{6}{c}{All 20 Languages} \\
            \multirow{2}{*}{Model Variant} & \multicolumn{3}{c|}{Hits@1} & \multicolumn{3}{c|}{MRR} & \multicolumn{3}{c|}{Hits@1} & \multicolumn{3}{c}{MRR} \\
             & Overall & Present & Missing & Overall & Present & Missing & Overall & Present & Missing & Overall & Present & Missing \\\hline
            simple fine-tuning & 0.443 & 0.860 & 0.287 & 0.522 & 0.888 & 0.385 & 0.584 & \textbf{0.883} & 0.350 & 0.649 & \textbf{0.907} & 0.451 \\
            \hspace{4mm}+dynamic ctxt removal (w/o neg) & 0.440 & 0.805 & 0.304 & 0.532 & 0.842 & 0.415 & 0.541 & 0.782 & 0.372 & 0.626 & 0.828 & 0.487$^{\dagger}$ \\
            \hspace{4mm}+dynamic ctxt removal & 0.473 & 0.846 & 0.334 & 0.547 & 0.875 & 0.424 & 0.574$^{\dagger}$ & 0.838$^{\dagger}$ & 0.376 & 0.649$^{\dagger}$ & 0.873$^{\dagger}$ & 0.486$^{\dagger}$ \\
            \hspace{4mm}+expansion & 0.648$^{\dagger}$ & 0.875 & 0.563$^{\dagger}$ & 0.719$^{\dagger}$ & \textbf{0.902} & 0.651$^{\dagger}$ & 0.657$^{\dagger}$ & 0.850 & 0.500$^{\dagger}$ & 0.733$^{\dagger}$ & 0.889 & 0.609$^{\dagger}$ \\
            \hspace{4mm}+knowledge injection & \textbf{0.677} & \textbf{0.879} & \textbf{0.602} & \textbf{0.741} & \textbf{0.902} & \textbf{0.681} & \textbf{0.672} & 0.877$^{\dagger}$ & \textbf{0.509} & \textbf{0.744} & 0.906 & \textbf{0.617} \\\hline
            \end{tabular}
        }
    \begin{tablenotes}
        \small \item[\dag] Indicates statistical significance ($p<0.05$) between the variant and the previous variant.
    \end{tablenotes}
    \end{threeparttable}
    \moveup
    \moveup
    \moveup
\end{table*}

First, we compare the performance of \xlokismall with PRP-Allpair, the state-of-the-art framework for ranking tasks using LLMs (Table~\ref{tab:zero_shot}). 
We find that \xlokismall substantially outperforms PRP-Allpair, both when using GPT-3.5 and GPT-4, particularly for the cases when the mention that is linked is not yet present in the text (\texttt{missing}).

Second, we evaluate our models on held-out data of the remaining 85 languages in Table~\ref{tab:language_scaling}. We reproduce a high zero-shot performance of \xlokismall, in comparison to results in the $9$ languages considered in Table~\ref{tab:zero_shot}. In comparison, other baseline models yield substantially lower performance.

Overall, these findings show that \xloki is capable of transferring the knowledge acquired during fine-tuning to unseen languages while maintaining a similar level of performance. 
This demonstrates that our entity insertion model can be scaled to many languages even if little or no additional training data is available for those languages. 

\subsection{Ablation analysis}
Finally, we investigate in more detail the effect of the extensions, namely, data augmentation, knowledge injection, and two-stage training that we introduce in the training pipeline of our model in comparison to a standard fine-tuning approach. 
Table~\ref{tab:novelties_analysis} portrays the improvement in performance on account for each extension introduced in this work.

Overall, we see that each extension has an overall positive impact on performance. 
First, introducing the dynamic context removal for data augmentation is only effective when including negative examples.
In that case, it improves the performance on the \texttt{missing} cases, but at the cost of performance in the \texttt{present} case. 
This is expected because context removal leads to the model seeing fewer training samples in the \texttt{present} case. 
Second, introducing expansion as a second stage in the training led to a large boost in performance in all scenarios, showing the benefit of using the smaller but high-quality dataset of added links for the training.
Third, the knowledge injection further improved the performance in both scenarios, indicating that the additional knowledge helps the model produce better relevance embeddings.

\section{Discussions}

\subsection{Summary of findings}
We introduced the novel task of entity insertion in information networks. 
Considering the case of Wikipedia, we justified the relevance and need for solving this task by demonstrating empirically that existing methods such as entity linking are often not suitable in practice.
In fact, we showed that in 65\% of edits in which links were inserted by editors, none of the existing text is suitable to insert the entity, i.e. new text has to be inserted \textit{somewhere in the article} along with the inserted entity. 

We developed a multilingual model (\xloki) to effectively solve the entity insertion task across 20 Wikipedia languages outperforming all other models.
First, our model substantially outperforms strong baseline approaches based on string matching or BM25, especially in the case when the linked mention was missing. We demonstrate how each of the introduced novelties (data augmentation, knowledge injection, two-stage training pipeline) contribute to improve the downstream performance.
Second, the multilingual model yields consistently better results than language-specific models. This shows that our model is capable of collating the knowledge acquired from each language to improve performance over all languages.
Third, our model works well in a zero-shot scenario, i.e. not only retaining over $95\%$ of the hypothetical best performance if the language was included but even outperforming the much larger GPT-3.5 and GPT-4. 
This demonstrates that the model is capable of transferring knowledge to languages unseen during fine-tuning which is crucial for the practical application across the more than 300 languages in Wikipedia, for which often there is little or no training data available. 
We compiled a new benchmark dataset for entity insertion in Wikipedia covering 105 languages. 
We make the dataset publicly available to enable future research in entity insertion. 

\subsection{Implications and broader impact}
\xhdr{A new benchmark for NLP tasks}
The problems of link recommendations and entity linking have been well-studied and many excellent solutions have been brought forward, some of which are denoted even near-optimal~\cite{Ghasemian2020stacking}.
The problem of entity insertion constitutes a new relevant and challenging task in the domain of NLP. 
Our multilingual dataset provides a resource for researchers for development and evaluation of new models to solve this task. 
This will help improve the overall capabilities of large language models when applied in the context of networks that are crucial for organizing textual information.

\xhdr{Supporting editors to bridge knowledge gaps}
Many articles in Wikipedia lack visibility in the hyperlink network capturing a specific aspect of the general problem of knowledge gaps~\cite{redi_taxonomy_2020}. 
For example, there are more than 8.8M so-called orphan articles~\cite{orphans}, \ie, articles without any incoming links, which are de-facto invisible to readers navigating Wikipedia. 
Even if suitable link targets are identified, a remaining challenge for editors is to identify a relevant position in the text where the link can be inserted.
At the current rate of ``de-orphanization'', it would take editors more than 20 years to work through the backlog of orphan articles, suggesting that existing tools do not support editors in addressing this issue effectively.
Our model can support editors in this task, complementing existing approached based on entity linking such as the add-a-link tool for newcomer editors~\cite{entity_linking_1}.

\section*{Limitations}

We tried different pre-trained language models for our experiments with RoBERTa outperforming BERT and T5 by a large margin. 
The use of larger models with more parameters could further improve performance.
While differences between RoBERTa-base and -large in English were marginal, we noticed a substantial drop when using the multilingual XLM-RoBERTa instead RoBERTa. 
This suggests that larger model architectures could be especially beneficial in the multilingual setting in order to improve support for low-resource languages, where performance is typically lower in comparison~\cite{wu_are_2020}. 
While multilingual models based on transformer architectures support many languages (e.g., XLM-RoBERTa was pre-trained on 100 languages), many of the more than 300 languages in Wikipedia are still not explicitly represented in the training data of these models. Thus, if unaddressed, the use of such models could lead to a language gap
constituting a substantial barrier towards knowledge equity~\cite{redi_taxonomy_2020}.

One practical limitation of the model is that the ranking of all text spans can become expensive if the article is very long and, thus, contains many candidates. 
This constitutes challenge for deploying the model in the future as a ready-to-use-tools for editors in practice.
This requires the integration of potential solutions for improving inference such as via hierarchical searching.

Further improvements to the model could come from integrating of additional information from the local Wikipedia graph structure or the candidate context. 
For example, a very strong signal are the links already existing in the candidate context, as these indicate entities related to the context. 
Providing these as additional features to the model might help generate better representations of the candidate~\cite{wikinav} and, as a result, better relevance embeddings. 
Furthermore, one could take advantage of the multilingual nature of Wikipedia with more than 300 language versions, each having a surprising amount of information not contained in any other languages~\cite{bao_omnipedia_2012}. 
Thus, existing content about a target entity from other languages could provide relevant context~\cite{lrec_candgen}, which could be made available through automatic translation, such as the already available section translation tool in Wikipedia~\cite{section_translation}.

In our operationalization of entity insertion, we assume that the link to be inserted consisting of the pair of the source- and target entity is known.
This assumption holds in the specific use-case of article ``de-orphanization''~\cite{orphans} serving as the motivation for formulating the task of entity insertion. 
However, when this is not the case, our model requires an additional step to generate a specific link, e.g., via existing link recommendation models. 

Our modeling framework is not suitable for the scenario where links are added in a section that did not exist in the previous version of the article (\texttt{missing\_section}). 
The text from the surrounding sections are not a good indicator for the insertion of a new entity, because they typically cover different subjects. 
The \texttt{missing\_section} scenario could be addressed through complementary approaches based on generative models that produce a draft for new section when none of the existing candidates leads to a high relevance score.

\section*{Ethics statement}
We have assessed the ethics of conducting this research and the output from the research and we have not identified ethical risks associated with the research. 
All the datasets and resources used in this work are publicly available and do not contain private or sensitive information about Wikipedia readers. 
While the dataset comes from individual edits (e.g., links added to Wikipedia articles), it does not contain any details about the author(s) of those edits.
All the findings are based on analyses conducted at an aggregate level, and thus, no individual-level inferences can be drawn. And lastly, the experiments are done on data already collected and no human subject has been involved as part of them. We confirm that we have read and abide by the ACL code of conduct.

\section*{Acknowledgements}
We thank Leila Zia, Alberto García-Durán, and Marija Šakota for insightful discussions and for reviewing an initial draft of this paper. West’s lab is partly supported by grants from the Swiss National Science Foundation (200021\_185043 and 211379), Swiss Data Science Center (P22\_08), H2020 (952215), Microsoft Swiss JRC, and Google. We also gratefully acknowledge generous gifts from Facebook, Google, and Microsoft.

{
\bibliography{locei}
\bibliographystyle{acl_natbib}
}
\pagebreak
\appendix

\section{Additional related work}
\label{app.additional_related_work}
\subsection{Pre-trained language models}
The transformer architecture, introduced by \cite{transformer}, has become the \textit{de facto} architecture for most Natural Language Processing (NLP) applications. A transformer-based pre-trained language model takes as input a text sequence and computes a vector embedding that captures the semantic and structural information contained in the text sequence, which can then be used in downstream applications.

Pre-training is an expensive process. For example, the base variant of BERT \citep{bert} took four days to train with 16 TPUs and RoBERTa \citep{roberta} took one day to train with 1024 GPUs. However, pre-trained models can be leveraged to novel downstream tasks by fine-tuning them on task-specific datasets. As a comparison, the authors of BERT \citep{bert} introduced several fine-tuned variants of BERT, all of which were fine-tuned in one hour using one TPU, which is much cheaper than pre-training the model for each task. This paradigm of pre-training language models on large amounts of data and then fine-tuning on much smaller amounts can reduce the cost of model training while retaining the knowledge from the pre-trained model and transferring it to the downstream task. Popular pre-trained models for multilingual tasks are mBERT \citep{bert}, XLM-RoBERTa \citep{xlm_roberta}, and mT5 \citep{mt5}.

\subsection{Ranking tasks}
Since entity insertion is a ranking task, in this section, we provide a short review of literature focusing on document retrieval and ranking. 

Classical approaches for ranking tasks, such as BM25 \cite{bm25}, mainly rely on probabilistic methods that attempt to match keywords between a query and a candidate document. However, these methods cannot capture complex semantic and structural patterns. For example, the sentences ``The hero defeated the dragon and saved the damsel'' and ``The knight slayed the beast and rescued the princess'' are semantically equivalent, but classical methods would to match them due to the small vocabulary overlap.

That said, pre-trained language models have become state-of-the-art for text ranking \cite{rank_llm}. A popular design for transformer-based ranking tasks is the cross-attention model, in which the query and the candidate document are concatenated into a sequence and then processed by the model. Since transformer models employ attention mechanisms, this strategy allows the model to capture the interactions between the query and the document.

This approach has been explored for encoder-only models \cite{encoder_rank_1, encoder_rank_2, encoder_rank_3}, outperforming classical methods. There has also been previous research \cite{encoder_decoder_rank_1, encoder_decoder_rank_2} in exploring encoder-decoder models, such as T5 \cite{t5}. However, even though encoder-decoder models are typically larger than encoder-only models, RankT5 \cite{rank_t5} has shown that there is no consistent winner between encoder-decoder and encoder-only models.
    
Given its recent success in document retrieval, the training objective of \loki is inspired by the ranking loss proposed in RankT5~\cite{rank_t5}.

\subsection{Domain adaption}

\cite{two_stage_background} have shown that a second phase of pre-training using domain-specific knowledge can improve the performance of language models. Their experiments started with a pre-trained RoBERTa model and continued pre-training it using unlabelled data from a large corpus of domain-specific text.

In our work, we propose a similar approach for fine-tuning, where we apply a first stage of domain-shifted data and then a second stage of domain-specific data to improve the performance further.

\section{Additional dataset processing details}
\label{app.data}

\subsection{Data preparation steps}
\label{app.data_process}

\xhdr{Existing links} For the existing links, we store the following data: source and target titles, Wikidata QIDs, lead paragraphs, the name of the section containing the link, and a context surrounding the link. The context is defined as the sentence containing the link and the five sentences before and after (or until we reach the end of the section). We additionally keep positional information about the mention and the sentence containing the mention relative to the context (\ie, the start and end indices of the mention and the sentence in the context). The positional information is relevant to the data augmentation strategy we introduced (see \S~\ref{app.dynamic_context_removal}). 

\xhdr{Added links} For the added links, we store the same information as in the existing links, except for the positional information. This is because positional information is required primarily for performing data augmentations, which are required only for processing existing links.

\subsection{Dataset statistics}
\label{app.data_statistics}

Table~\ref{tab:data_statistics} shows the summary statistics of the entity insertion dataset for each of the 105 considered language versions of Wikipedia, in particular the number of articles, the number of existing links, and the number of added links.
\begin{table*}[!htb]
    \centering
    \caption{Summary statistics of the full entity insertion dataset collected from 105 different Wikipedia language versions.}
    \resizebox{0.99\textwidth}{!}{%
    \begin{tabular}{ccccc|ccccc}
     & Language  & Articles & Existing Links & Added Links & & Language & Articles & Existing Links & Added Links \\\hline
en & English & 6.7M & 166M & 368K & de & German & 2.8M & 78.3M & 94.3K\\
sv & Swedish & 2.5M & 29.9M & 10.7K & fr & French & 2.5M & 85.1M & 64.5K\\
nl & Dutch & 2.1M & 24.7M & 23.6K & ru & Russian & 1.9M & 47.6M & 33.8K\\
es & Spanish & 1.8M & 47.9M & 66.3K & it & Italian & 1.7M & 51.1M & 45.6K\\
pl & Polish & 1.5M & 30.1M & 27.2K & ja & Japanese & 1.3M & 60.6M & 79.0K\\
zh & Chinese & 1.3M & 23.1M & 28.2K & vi & Vietnamese & 1.2M & 10.3M & 11.9K\\
ar & Arabic & 1.2M & 16.3M & 17.8K & pt & Portuguese & 1.1M & 21.9M & 24.2K\\
fa & Persian & 971K & 9.5M & 18.1K & ca & Catalan & 732K & 14.6M & 18.4K\\
sr & Serbian & 671K & 8.3M & 5.4K & id & Indonesian & 650K & 8.5M & 13.7K\\
ko & Korean & 634K & 11.2M & 21.3K & no & Norwegian & 611K & 11.3M & 7.2K\\
ce & Chechen & 599K & 3.0M & 48 & fi & Finnish & 554K & 9.7M & 13.7K\\
cs & Czech & 531K & 14.4M & 12.3K & tr & Turkish & 531K & 6.7M & 14.9K\\
hu & Hungarian & 527K & 10.6M & 7.8K & tt & Tatar & 496K & 3.1M & 94\\
sh & Serbo-Croatian & 456K & 8.3M & 807 & ro & Romanian & 439K & 6.9M & 4.2K\\
eu & Basque & 412K & 4.4M & 5.1K & ms & Malay & 363K & 2.9M & 2.7K\\
he & Hebrew & 341K & 14.7M & 36.7K & eo & Esperanto & 340K & 6.7M & 5.8K\\
hy & Armenian & 296K & 4.5M & 3.7K & da & Danish & 294K & 5.7M & 2.3K\\
bg & Bulgarian & 288K & 5.2M & 4.8K & cy & Welsh & 270K & 2.6M & 386\\
sk & Slovak & 242K & 3.4M & 3.1K & azb & South Azerbaijani & 242K & 1.0M & 22\\
simple & Simple English & 240K & 2.6M & 3.8K & et & Estonian & 235K & 4.4M & 4.7K\\
kk & Kazakh & 233K & 1.6M & 1.5K & be & Belarusian & 232K & 3.1M & 3.0K\\
uz & Uzbek & 230K & 1.3M & 4.3K & min & Minangkabau & 226K & 644K & 21\\
el & Greek & 224K & 4.8M & 6.7K & lt & Lithuanian & 210K & 3.8M & 2.4K\\
gl & Galician & 196K & 3.9M & 4.2K & hr & Croatian & 194K & 3.2M & 3.4K\\
ur & Urdu & 190K & 1.4M & 5.2K & az & Azerbaijani & 188K & 2.4M & 6.3K\\
sl & Slovenian & 182K & 3.1M & 1.5K & ka & Georgian & 163K & 2.3M & 1.7K\\
ta & Tamil & 157K & 1.6M & 1.1K & hi & Hindi & 157K & 1.2M & 2.3K\\
la & Latin & 138K & 2.5M & 1.2K & mk & Macedonian & 136K & 2.3M & 923\\
ast & Asturian & 128K & 2.6M & 49 & lv & Latvian & 121K & 2.0M & 1.9K\\
af & Afrikaans & 111K & 1.3M & 1.4K & tg & Tajik & 108K & 567K & 123\\
sq & Albanian & 97.3K & 873K & 523 & mg & Malagasy & 95.7K & 495K & 1.2K\\
bs & Bosnian & 89.7K & 1.6M & 969 & oc & Occitan & 88.3K & 1.1M & 1.9K\\
te & Telugu & 82.2K & 934K & 1.1K & sw & Swahili & 74.3K & 1.0M & 558\\
lmo & Lombard & 71.9K & 380K & 26 & jv & Javanese & 70.5K & 513K & 161\\
ba & Bashkir & 62.4K & 960K & 649 & lb & Luxembourgish & 61.7K & 930K & 754\\
mr & Marathi & 60.9K & 409K & 67 & su & Sundanese & 60.3K & 470K & 6\\
is & Icelandic & 56.4K & 725K & 1.0K & ga & Irish & 56.0K & 387K & 204\\
ku & Kurdish & 54.3K & 252K & 614 & fy & Western Frisian & 51.0K & 1.3M & 579\\
pa & Punjabi & 49.4K & 282K & 139 & cv & Chuvash & 48.3K & 213K & 304\\
br & Breton & 46.5K & 326K & 852 & tl & Tagalog & 43.2K & 435K & 512\\
an & Aragonese & 40.8K & 620K & 70 & io & Ido & 40.7K & 422K & 230\\
sco & Scots & 35.5K & 251K & 40 & vo & Volapük & 34.6K & 134K & 7\\
ne & Nepali & 32.1K & 168K & 250 & ha & Hausa & 30.6K & 129K & 262\\
gu & Gujarati & 30.2K & 411K & 29 & kn & Kannada & 28.0K & 253K & 514\\
bar & Bavarian & 27.0K & 207K & 21 & scn & Sicilian & 23.8K & 132K & 5\\
mn & Mongolian & 22.5K & 187K & 467 & si & Sinhala & 20.3K & 81.7K & 36\\
ps & Pashto & 16.2K & 49.7K & 10 & gd & Scottish Gaelic & 15.8K & 207K & 14\\
yi & Yiddish & 15.2K & 185K & 21 & sd & Sindhi & 13.4K & 49.5K & 14\\
am & Amharic & 12.9K & 69.1K & 12 & as & Assamese & 11.9K & 104K & 459\\
sa & Sanskrit & 10.5K & 65.2K & 18 & km & Khmer & 9.8K & 52.3K & 95\\
ary & Moroccan Arabic & 8.0K & 50.5K & 129 & so & Somali & 7.4K & 64.2K & 60\\
ug & Uyghur & 5.9K & 9.7K & 1 & lo & Lao & 4.7K & 14.2K & 11\\
om & Oromo & 1.7K & 5.0K & 18 & xh & Xhosa & 1.6K & 2.8K & 1\\\hline
    \end{tabular}
    }
    \label{tab:data_statistics}
\end{table*}

Table~\ref{tab:data_statistics_train_test} shows the number of samples contained in the training and test splits, respectively, for each of the 20 Wikipedia language versions considered in the experiments.
\begin{table*}[!htb]
    \centering
    \caption{Summary statistics of the train and test sets for 20 Wikipedia language versions considered in the experiments.}
    \resizebox{0.7\textwidth}{!}{%
    \begin{tabular}{cc|cccc}
         & \multirow{2}{*}{Language}  & \multirow{2}{*}{Articles} & \multirow{2}{*}{Existing Links} & \multicolumn{2}{c}{Added Links} \\
         \cline{5-6}
         & & &  & Train & Test \\\hline
        en     & English & 6.7M & 166M & 552K & 416K \\
        fr     & French & 2.5M & 85M & 130K & 76K \\
        it     & Italian & 1.8M & 51M & 101K & 56K \\
        ja     & Japanese & 1.4M & 61M & 150K & 111K \\
        pt     & Portuguese & 1.1M & 22M & 54K & 32K \\
        cs     & Czech & 526K & 14M & 27K & 15K \\
        ms     & Malay & 362K & 2.9M & 6K & 3K \\
        cy     & Welsh & 269K & 2.7M & 1K & 455 \\
        sk     & Slovak & 240K & 3.4M & 7K & 4.3K \\
        simple & Simple English & 238K & 2.6M & 9.4K & 4.8K \\
        kk     & Kazakh & 232K & 1.6M & 2.7K & 2.0K \\
        uz     & Uzbek & 224K & 1.3M & 12K & 5.9K \\
        ur     & Urdu & 188K & 1.4M & 14K & 7.5K \\
        hi     & Hindi & 155K & 1.2M & 3.2K & 3.2K \\
        af     & Afrikaans & 111K & 1.4M & 3.3K & 1.7K \\
        sw     & Swahili & 73K & 1.0M & 1.1K & 616 \\
        ga     & Irish& 56K & 380K & 849 & 256 \\
        is     & Icelandic & 51K & 610K & 1.6K & 1.2K \\
        gu     & Gujarati & 30K & 410K & 197 & 48 \\
        kn     & Kannada & 27K & 250K & 1.1K & 609 \\\hline
    \end{tabular}
    }
    \label{tab:data_statistics_train_test}
\end{table*}

\subsection{Entity insertion categories}
\label{app.link_insertion_examples}
Table~\ref{tab:link_insertion_examples} shows an example for each of the entity insertion categories, except for the category \texttt{missing\_section}, demonstrating that the problem of entity insertion grows in complexity as more text is missing. 
\begin{table*}[!htb]
    \centering
    \caption{Examples of different entity insertion categories observed when adding links in Wikipedia. The added link is marked in \textcolor{blue}{blue}.}
    {\renewcommand{\arraystretch}{1.5}
    \begin{tabular}{c|p{2.35in}p{2.35in}}
    Strategy & First Version Text & Second Version Text \\\hline
    Text Present & It is best eaten when it is somewhat below normal room temperature. In most countries, brie-style cheeses are made with Pasteurized milk. & It is best eaten when it is somewhat below normal \textcolor{blue}{room temperature}. In most countries, brie-style cheeses are made with Pasteurized milk. \\\hline
    Missing Mention & Vercetti Regular, also known as Vercetti, is a free font that can be used for both commercial and personal purposes. It became available in 2022 under the Licence Amicale, which allows users to share the font files with friends and colleagues. & Vercetti Regular, also known as Vercetti, is a free font (\textcolor{blue}{freeware}) that can be used for both commercial and personal purposes. It became available in 2022 under the Licence Amicale, which allows users to share the font files with friends and colleagues. \\\hline
    Missing Sentence & \textcolor{white}{Kivi was born in Nurmijärvi.} Kivi lived in time when all educated people in Finland spoke Swedish. He was the first professional writer who published his works in Finnish. Kivi, Mikael Agricola and Elias Lönnrot are regarded fathers of a national literature in Finnish. & Kivi was born in \textcolor{blue}{Nurmijärvi}. He lived in time when all educated people in Finland spoke Swedish. He was the first professional writer who published his works in Finnish. Kivi, Mikael Agricola and Elias Lönnrot are regarded fathers of a national literature in Finnish. \\\hline
    Missing Span & The game will be released for Windows PC, Mac and Linux, with Nintendo Switch being the only console to receive the game at launch. \newline \textcolor{white}{During the Xbox \& Bethesda Games Showcase, it was revealed that the game would be coming to Xbox Game Pass through PC and Xbox Series X/S. It was also revealed that the game would be coming to PlayStation 4 and PlayStation 5.} \newline Originally, Hornet was planned as a second playable character to be included in a downloadable content pack (DLC) for Hollow Knight, funded as a stretch goal in the game's Kickstarter campaign. & The game will be released for Windows PC, Mac and Linux, with Nintendo Switch being the only console to receive the game at launch. \newline During the Xbox \& Bethesda Games Showcase, it was revealed that the game would be coming to Xbox Game Pass through PC and Xbox Series X/S. It was also revealed that the game would be coming to PlayStation 4 and \textcolor{blue}{PlayStation 5}. \newline Originally, Hornet was planned as a second playable character to be included in a downloadable content pack (DLC) for Hollow Knight, funded as a stretch goal in the game's Kickstarter campaign.\\\hline
    \end{tabular}
    }
    \label{tab:link_insertion_examples}
\end{table*}

\begin{figure*}[!htb]
    \centering
    \includegraphics[width=\textwidth]{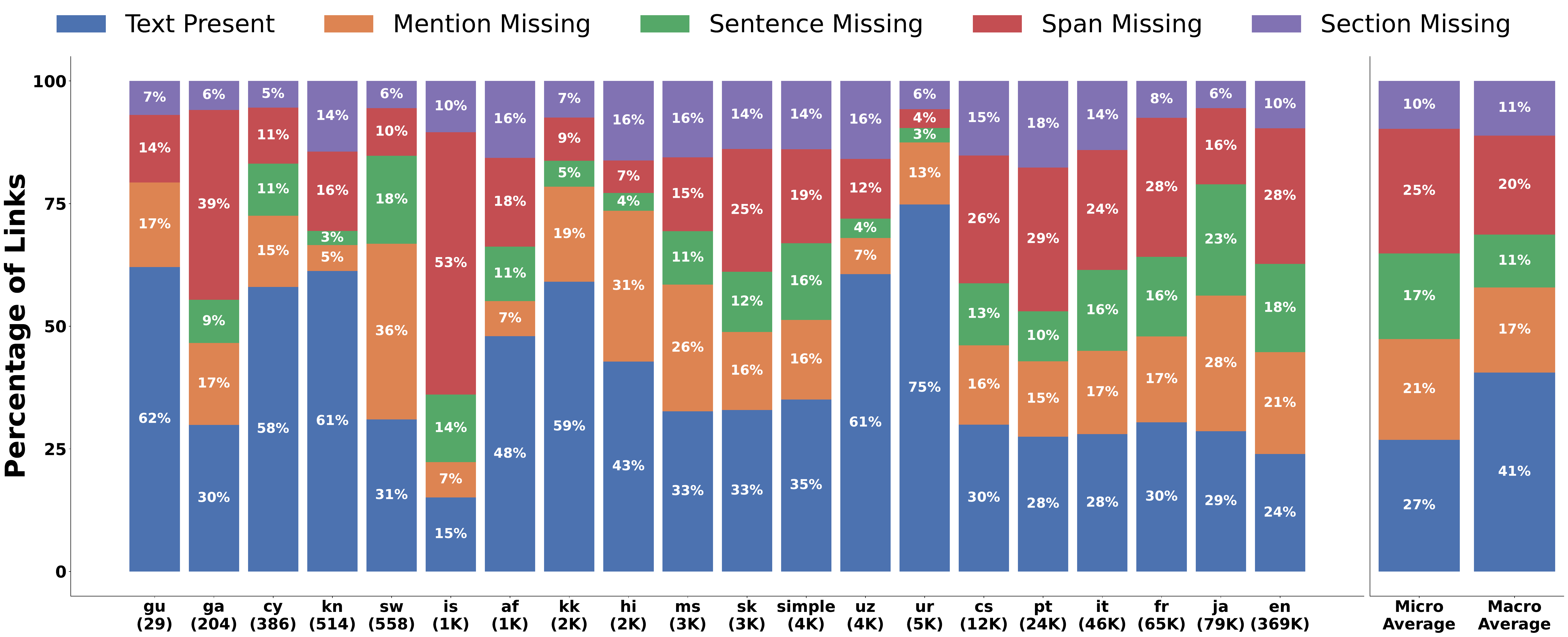}
    \caption{The distribution of entity insertion categories across the 20 considered Wikipedia language versions from October to November 2023. The x-axis shows the language code and the number of links added in each language.}
    \label{fig:full_link_distributions}
\end{figure*}

Additionally, Fig.~\ref{fig:full_link_distributions} shows the distribution of entity insertion categories for 20 Wikipedia language versions considered in the experiments.

\subsection{Dynamic context removal}
\label{app.dynamic_context_removal}

Table~\ref{tab:context_mask} shows examples of the different types of dynamic context removal. Specifically, we randomly remove a word (\texttt{rm\_mention} simulation), a sentence (\texttt{rm\_sent} simulation), or a span of sentences (\texttt{rm\_span} simulation) during training. Before sending the input to the model, we randomly select one of the masking strategies mentioned above (as well as no masking) to modify the input accordingly. However, before applying the strategy, we verify if the selected strategy does not produce an empty input. This may happen when, for example, the context is a single sentence, in which case simulating the \texttt{rm\_sent} strategy would lead to an empty input. If the sampled strategy would produce an empty input, we re-sample a less aggressive strategy.

While performing the \texttt{rm\_span} simulation, the number of sentences to remove is chosen randomly between $2$ and $5$. Note that we used a space-based splitting for ease of implementation, and we acknowledge that this could be an issue for certain languages, such as Japanese or Chinese, which we intend to fix in the future.

\begin{table*}[!htb]
    \centering
    \caption{Examples of different strategies for dynamic context removal. The mention of the target link is marked in \textcolor{blue}{blue}.}
    {\renewcommand{\arraystretch}{1.5}
    \begin{tabular}{p{1.2in}|p{2.3in}p{2.3in}}
    Strategy & Original Text & Modified Text \\\hline
    No removal\newline(\texttt{rm\_nth}) & Pulaski County is a county located in the central portion of the U.S. state of Georgia. As of the 2020 census, the population was 9,855. The \textcolor{blue}{county seat} is Hawkinsville. & Pulaski County is a county located in the central portion of the U.S. state of Georgia. As of the 2020 census, the population was 9,855. The county seat is Hawkinsville. \\
    Mention removal\newline(\texttt{rm\_mention}) & Perthes-lès-Brienne is a commune of the Aube \textcolor{blue}{département} in the north-central part of France. & Perthes-lès-Brienne is a commune of the Aube ~~ in the north-central part of France. \\
    Sentence removal\newline(\texttt{rm\_sent}) & In this Japanese name, the family name is Fujita. Yoshiaki Fujita (born \textcolor{blue}{12 January} 1983) is a Japanese football player. He plays for Oita Trinita. & In this Japanese name, the family name is Fujita. \newline\newline ~~~~ He plays for Oita Trinita. \\
    Span removal\newline(\texttt{rm\_span}) & Administration\newline
The department of French Guiana is managed by the Collectivité territorial de la Guyane in Cayenne. There are 2 \textcolor{blue}{arrondissements} (districts) and 22 communes (municipalities) in French Guiana. The cantons of the department were eliminated on 31 December 2015 by the Law 2011-884 of 27 July 2011.\newline
The 22 communes in the department are: & Administration\newline
\newline
\newline
\newline
\newline
\newline
\newline
\newline
\newline
The 22 communes in the department are: \\\hline
    \end{tabular}
    }
    \label{tab:context_mask}
\end{table*}

\subsection{Rules for sampling negative candidates} 
\label{app.negative_sampling}

We employ the following rules when constructing the negative candidates, both for training and validation. 
\begin{enumerate}
    \item A candidate's context should not span over two different sections.
    \item A candidate's context should not contain any of the mentions previously used to link to the target entity. 
\end{enumerate}

The first rule keeps the content of each context consistent, as two distinct sections can cover very different topics. The second rule ensures that all the candidates used to evaluate the module are correctly classified as either positive candidates or negative candidates. For example, if the goal is to insert the entity ``1984'' (the book - \verb|Q208460|) and there is a sentence in the article with the word ``1984'' not linked to the target article, there could be three reasons for the link to be missing. First, the mention ``1984'' could be related to a different entity (\eg, the year - \verb|Q2432|), in which case the sentence should belong to a negative candidate. Second, the mention is supposed to be for the target entity but it is not yet linked, in which case the sentence should belong to an additional positive candidate. Finally, the mention is supposed to be for the target entity but it should not be linked because of Wikipedia's editing guidelines, in which case it is not clear whether the sentence should belong to a negative or a positive candidate. Due to this unclear categorization, we choose to remove any sentences containing mentions previously associated with the target entity to be inserted.

\section{Additional experiments}
\label{app:additional_experiments}

\subsection{Hyperparameters}
We train the encoder and MLP with learning rates of $1e-5$ and $1e-4$, respectively, using $N=9$ negative candidates. Moreover, we use $5$ sentences on either side as context for each candidate text span and set $|M_{tgt}|=10$. The first stage of training uses $20K$ data points and is trained for $4$ epochs, whereas the second stage uses all the available data for $2$ epochs. Mimicking the real-world entity insertion scenarios, we set \texttt{rm\_nth}=40\%, \texttt{rm\_mention}=20\%, \texttt{rm\_sentence}=30\%, and \texttt{rm\_span}=10\%.

\subsection{Multilingual entity insertion stratified by language}
\label{app.main_result_by_language} 

\begin{figure*}[!htb]
    \centering
    \includegraphics[width=\textwidth]{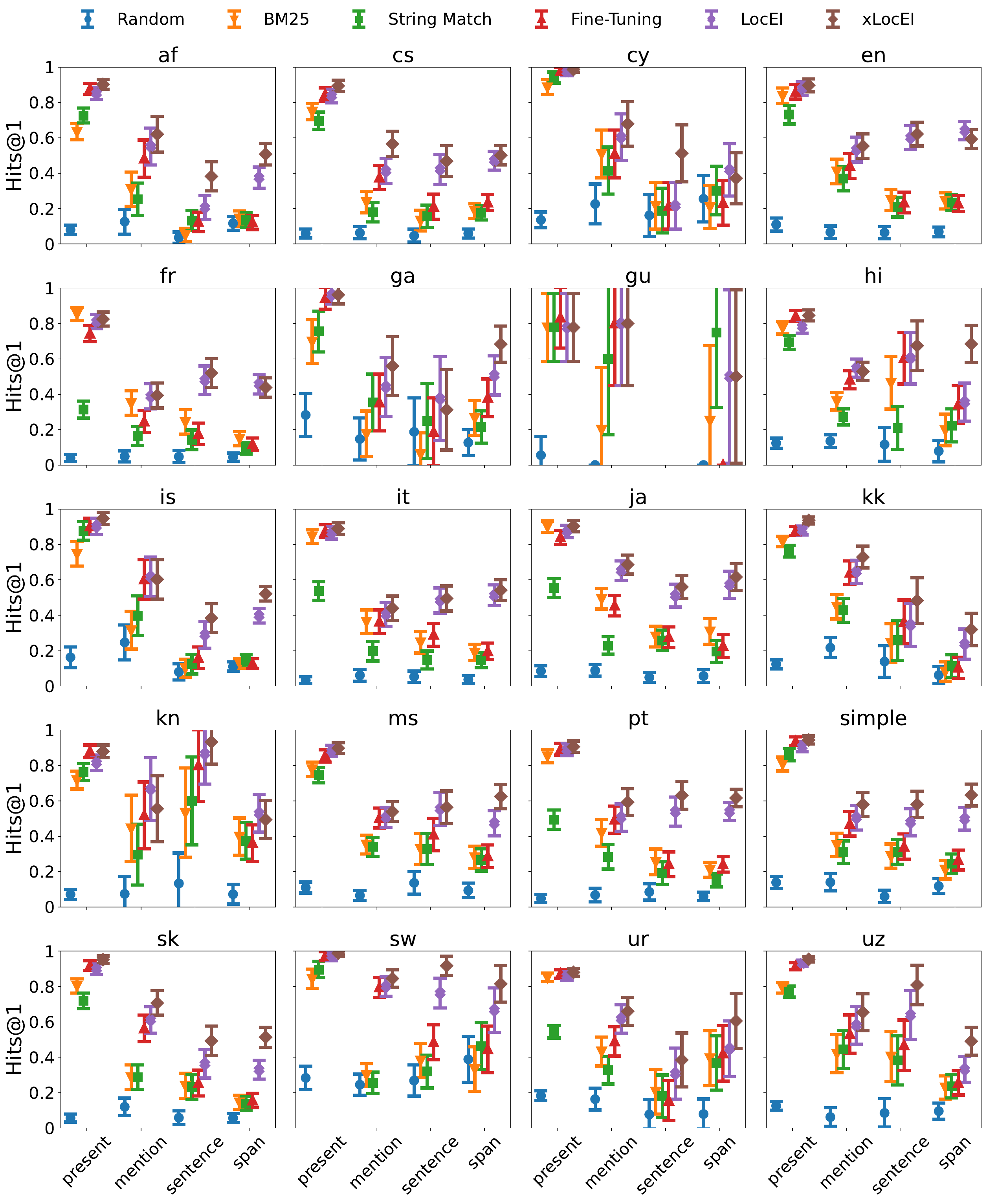}
    \caption{Entity insertion performance across all 20 Wikipedia language versions measured using hits@1. \xloki trains a single model jointly on all 20 languages, whereas other methods train a separate model for each language. The categorization of entity insertion types is discussed in \S~\ref{sec:data}.}
    \label{fig:full_hits_at_1}
\end{figure*}

\begin{figure*}[!htb]
    \centering
    \includegraphics[width=\textwidth]{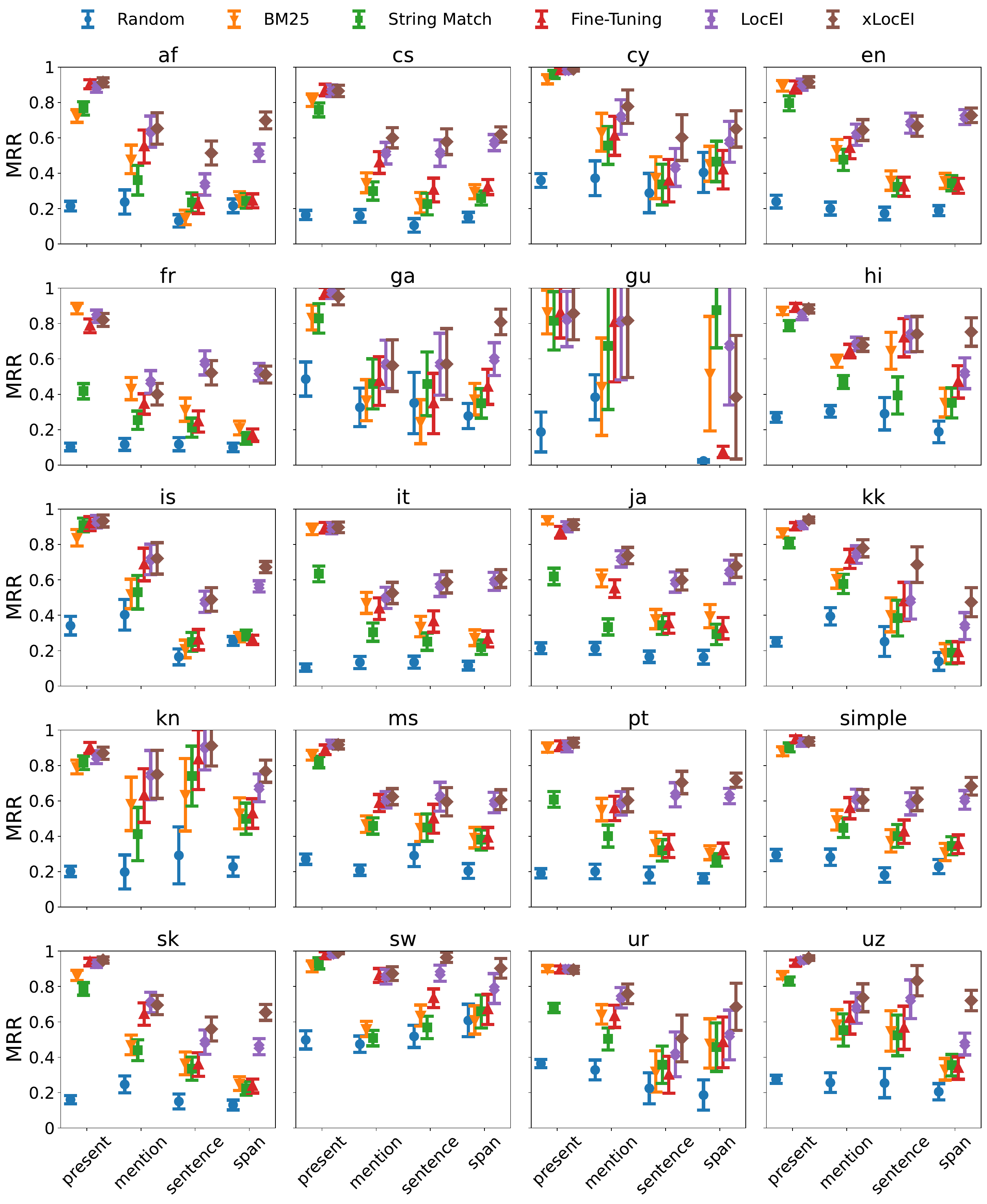}
    \caption{Entity insertion performance across all 20 Wikipedia language versions measured using MRR. \xloki trains a single model jointly on all 20 languages, whereas other methods train a separate model for each language. The categorization of entity insertion types is discussed in \S~\ref{sec:data}.}
    \label{fig:full_mrr}
\end{figure*}

Figs.~\ref{fig:full_hits_at_1} and~\ref{fig:full_mrr} portray the entity insertion performance stratified by language of all the benchmarked methods using hits@1 and MRR, respectively. \tomas{The results clearly show that, as entity insertion becomes more complex, the baselines start to decrease in performance, being significantly outperformed by \loki and \xloki.}

\subsection{Zero-shot entity insertion stratified by language}
\label{app.zeroshot_result_by_language}

Table~\ref{tab:zero_shot_models} provides additional details about the data such as the languages and the size of the datasets, used to train the different variants of the multilingual models employed in the zero-shot setting.

\begin{table*}[!htb]
    \centering
    \caption{Details about the languages and size of the dataset used to train the two \xloki model variants, \ie, \xlokilarge and \xlokismall.}
    \resizebox{0.98\textwidth}{!}{%
    \begin{tabular}{c|cp{2in}cc}
    \multirow{2}{*}{Model} & \multirow{2}{*}{Starting Model} & \multirow{2}{*}{Fine-Tuned Languages} & \multicolumn{2}{c}{Training Data Size} \\
     &  &  & Stage 1 & Stage 2 \\\hline
    xLocEI$_{20}$ & XLM-RoBERTa$_{\text{BASE}}$ & en, fr, it, ja, pt, cs, ms, cy, sk, uz, simple, kk, ur, hi, af, sw, ga, is, kn, gu & 20K & 503K \\
    xLocEI$_{11}$ & XLM-RoBERTa$_{\text{BASE}}$ & en, it, ja, cs, cy, uz, ur, hi, sw, is, kn & 20K & 348K \\\hline
    \end{tabular}
    }
    \label{tab:zero_shot_models}
\end{table*}

\begin{figure*}[!htb]
    \centering
    \includegraphics[width=\textwidth]{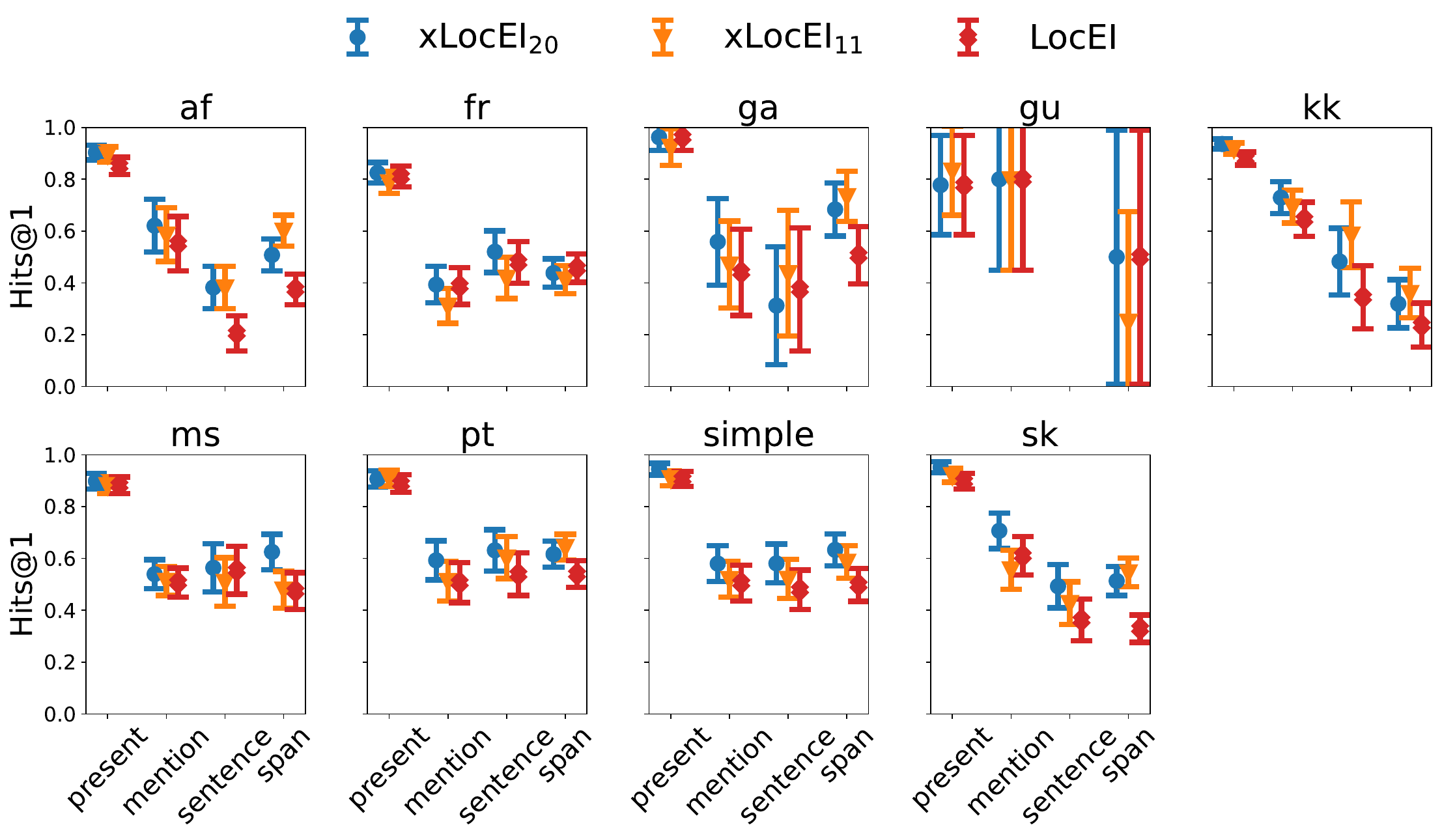}
    \caption{Entity insertion performance measured using hits@1 in the zero-shot setting: results across 9 Wikipedia language versions that were not used for fine-tuning \xlokismall. \xlokilarge was trained jointly on all 20 languages, whereas \loki trains a separate model for each language. The categorization of entity insertion types is discussed in \S~\ref{sec:data}.}
    \label{fig:zero_shot_hits}
\end{figure*}

\begin{figure*}[!htb]
    \centering
    \includegraphics[width=\textwidth]{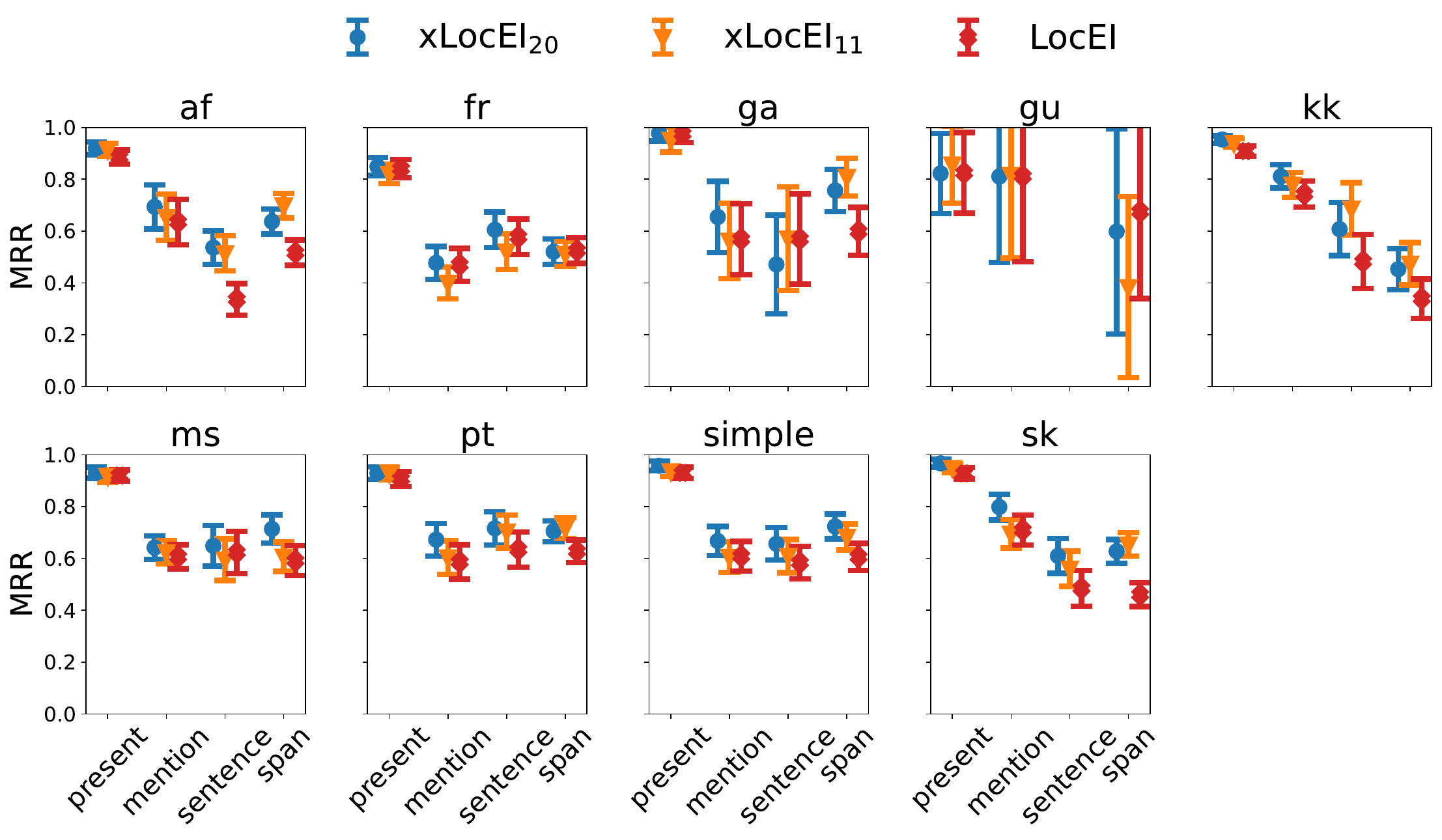}
    \caption{Entity insertion performance measured using MRR in the zero-shot setting: results across 9 Wikipedia language versions that were not used for fine-tuning \xlokismall. \xlokilarge was trained jointly on all 20 languages, whereas \loki trains a separate model for each language. The categorization of entity insertion types is discussed in \S~\ref{sec:data}.}
    \label{fig:zero_shot_mrr}
\end{figure*}

Figs.~\ref{fig:zero_shot_hits} and~\ref{fig:zero_shot_mrr} portray the zero-shot entity insertion performance stratified by language of all the benchmarked methods using hits@1 and MRR, respectively.

\subsection{Impact of the starting model}

Since our approach is based on fine-tuning pre-trained models, the starting pre-trained model may have an impact on the eventual model performance. We studied this dependence using three pre-trained models: BERT$_{\text{BASE}}$, RoBERTa$_{\text{BASE}}$ and the encoder portion of T5$_{\text{BASE}}$, which we call T5$_{\text{BASE}}^{\text{enc}}$. We considered BERT and RoBERTa because they are amongst the most popular transformer encoder models. We additionally included T5 to see how encoder-decoder models perform in the entity insertion task. However, as RankT5 \cite{rank_t5} showed there was no clear benefit in using the full encoder-decoder architecture, as opposed to encoder-only architecture, and thus, for computational reasons we decided to use the encoder-only variant of T5, T5$^{\text{enc}}$.

\begin{table*}[!htb]
    \centering
    \caption{Comparing the entity insertion performance obtained for Simple English with different starting models. The categorization of entity insertion types into `Overall', `Missing', and `Present' is discussed in \S~\ref{subsec:setup}.}
    \label{tab:starting_model}
    \begin{threeparttable}
        \begin{tabular}{c|ccc|ccc}
        \multirow{2}{*}{Method} & \multicolumn{3}{c|}{Hits@1} & \multicolumn{3}{c}{MRR} \\
         & Overall & Present & Missing & Overall & Present & Missing \\\hline
        BERT & 0.666 & 0.916 & 0.492 & 0.738 & 0.940 & 0.598 \\
        T5$^{\text{enc}}$ & 0.710 & 0.929 & 0.558 & 0.774 & 0.952 & 0.650 \\
        RoBERTa & \textbf{0.851}$^{\dagger}$ & \textbf{0.957}$^{\dagger}$ & \textbf{0.777}$^{\dagger}$ & \textbf{0.890}$^{\dagger}$ & \textbf{0.968} & \textbf{0.835}$^{\dagger}$ \\
        \hline
        \end{tabular}
        \begin{tablenotes}
            \small \item[\dag] Indicates statistical significance ($p<0.05$) between the best and the second-best scores.
        \end{tablenotes}
    \end{threeparttable}
\end{table*}

We trained each model on the Simple English dataset, and we measured their performance on the test data. Table~\ref{tab:starting_model} shows that the RoBERTa model outperformed both BERT and T5$^{\text{enc}}$ in all entity insertion categories by a large margin. BERT and T5$^{\text{enc}}$ performed similarly, with T5$^{\text{enc}}$ doing slightly better. These results may be explained by the fact that the RoBERTa tokenizer has a much larger vocabulary than the tokenizers for BERT or T5$^{\text{enc}}$. A larger vocabulary might make it possible for the model to capture more fine-grained linguistic and structural patterns in the candidate text spans, enabling the model to exploit patterns that neither T5$^{\text{enc}}$ nor BERT can capture. 

\subsection{Impact of the model size}
\label{sec:model_size}

There is a widely known trend in the deep learning community that bigger models tend to perform better than smaller models \cite{big_models_1, big_models_2, big_models_3}. To this end, we studied how the model size impacts the entity insertion performance by comparing RoBERTa$_{\text{LARGE}}$ with RoBERTa$_{\text{BASE}}$ on the Simple English dataset.

Table~\ref{tab:model_size} shows that there is no statistically significant difference between the performance of RoBERTa$_{\text{LARGE}}$ and RoBERTa$_{\text{BASE}}$. These results point to the fact that the increased model complexity is not sufficient to improve model performance. It is worth noting that these results were obtained for Simple English. The multilingual problem is much harder and it might benefit from the increased complexity and larger parameter space of the larger model. We leave this study for future work.

Additionally, these findings give more strength to the hypothesis that the reason why RoBERTa is significantly better than BERT and T5$^{\text{enc}}$ is because RoBERTa's larger tokenizer allows the model to capture more fine-grained linguistic and structural patterns in the candidate. This increased input representation space seems to be vital for entity insertion.

\begin{table*}[!htb]
    \centering
    \caption{Comparing the entity insertion performance obtained for Simple English with varying model sizes. The categorization of entity insertion types is discussed in \S~\ref{sec:data}.}
    \begin{tabular}{c|cccccccc}
    \multirow{2}{*}{Model} & \multicolumn{2}{c}{Text Present} & \multicolumn{2}{c}{Missing Mention} & \multicolumn{2}{c}{Missing Sentence} & \multicolumn{2}{c}{Missing Span} \\
     & Hits@1 & MRR & Hits@1 & MRR & Hits@1 & MRR & Hits@1 & MRR \\\hline
     RoBERTa$_{\text{BASE}}$ & 0.956 & 0.968 & \textbf{0.696} & \textbf{0.760} & 0.834 & 0.884 & 0.799 & 0.859 \\
     RoBERTa$_{\text{LARGE}}$ & \textbf{0.964} & \textbf{0.975} & 0.670 & 0.744 & \textbf{0.856} & \textbf{0.895} & \textbf{0.822} & \textbf{0.873} \\\hline
    \end{tabular}
    \label{tab:model_size}
\end{table*}

\subsection{Impact of the size of training data}
\label{sec:train_size}
As discussed in \S~\ref{subsec:augmentation}, we use the existing and added links data during the first and second stages of our training pipeline, respectively. In this analysis, we studied how much data is needed for each stage. To study the impact of the training data size on the downstream entity insertion performance, we trained a RoBERTa$_{\text{BASE}}$ model with varying portions of the full English dataset.

\begin{figure*}[!htb]
     \centering
         \includegraphics[width=.85\textwidth]{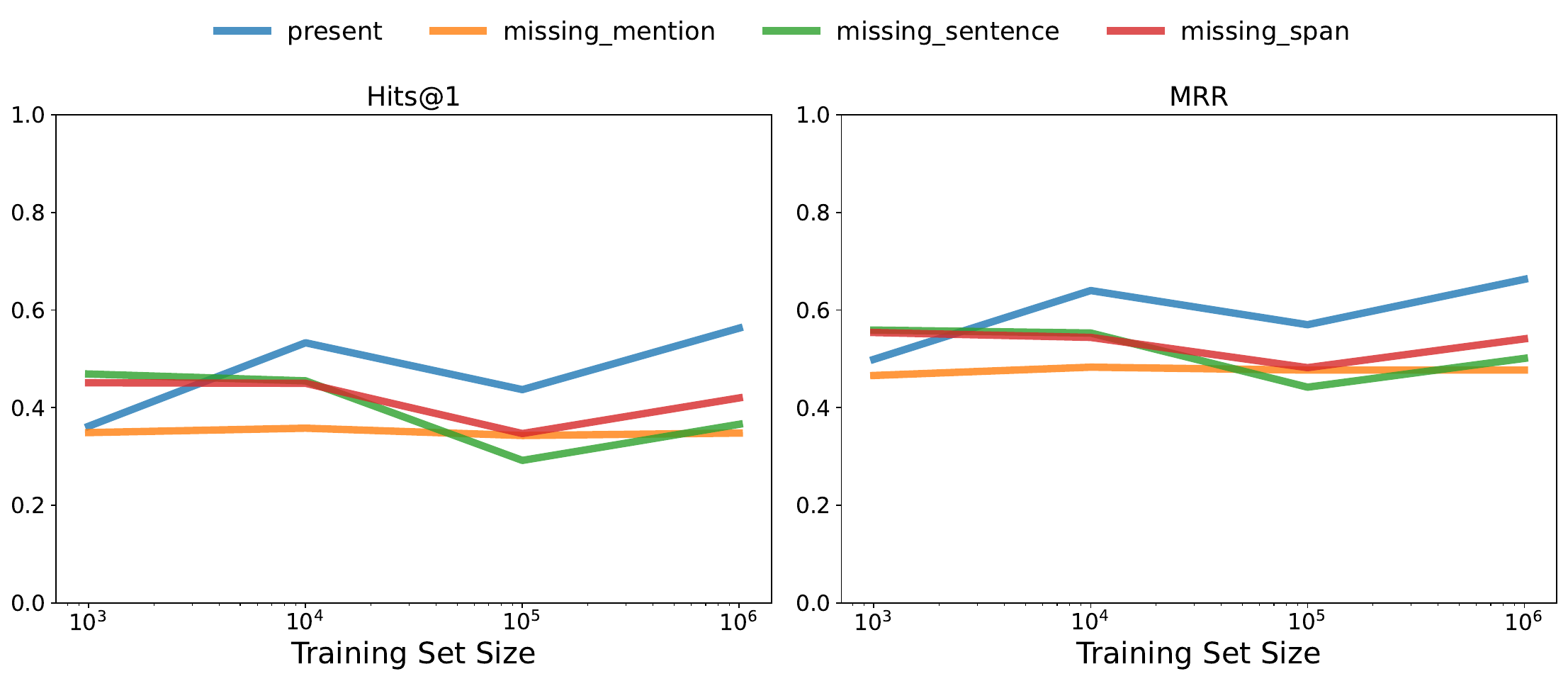}
         \caption{Impact of the amount of data used in the first stage on the downstream entity insertion performance. Note that the model is trained solely using the first stage. The categorization of entity insertion types is discussed in \S~\ref{sec:data}.}
         \label{fig:train_size_1}
\end{figure*}

\begin{figure*}[!htb]
\centering
         \includegraphics[width=.85\textwidth]{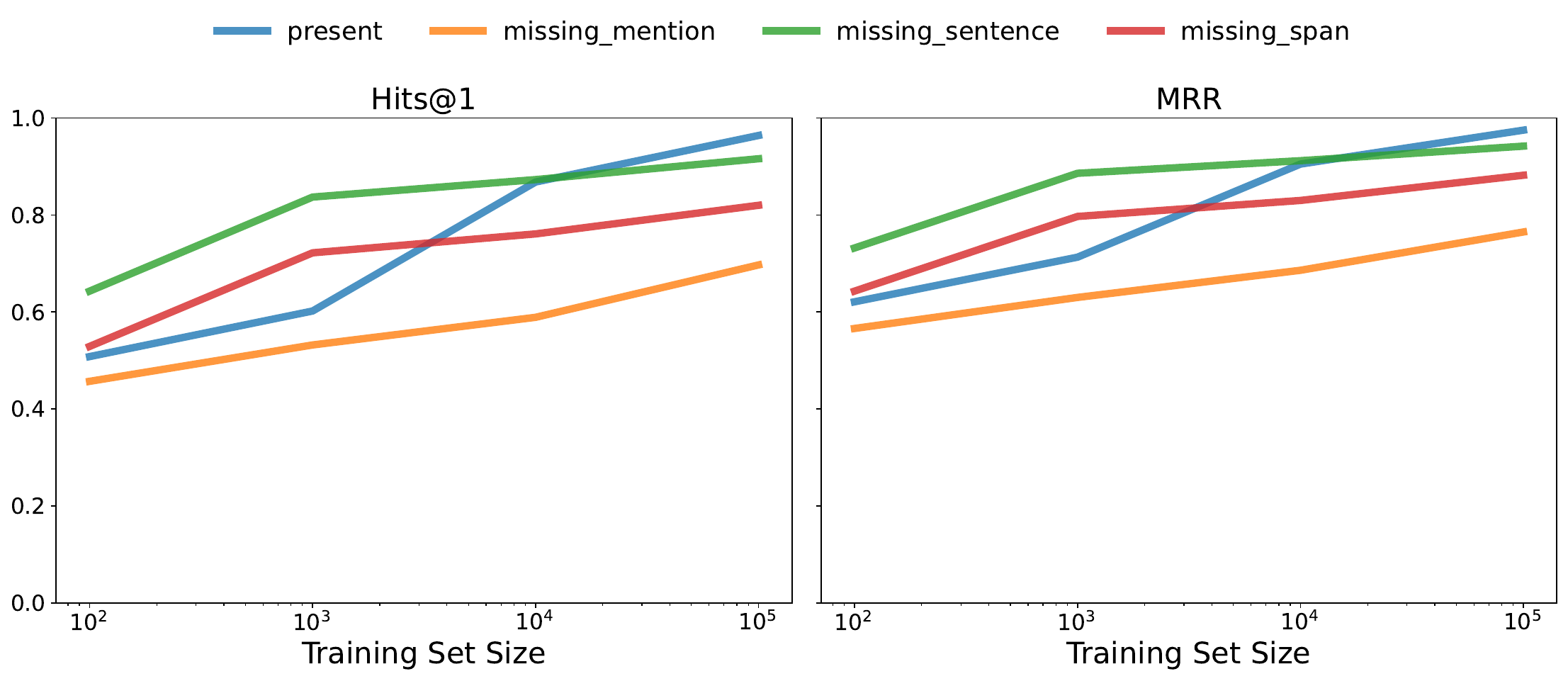}
         \caption{Impact of the amount of data used in the second stage on the downstream entity insertion performance. Note that the model is trained solely using the second stage. The categorization of entity insertion types is discussed in \S~\ref{sec:data}.}
         \label{fig:train_size_2}
\end{figure*}

Fig.~\ref{fig:train_size_1} shows the performance of \loki for different entity insertion categories with varying training data sizes og $\{10^3,10^4,10^5,10^6\}$ in the first stage of the training pipeline. Note that \loki was trained using only the first stage for this analysis. Fig.~\ref{fig:train_size_2} shows an analogous plot for the second stage with varying training data sizes of $\{10^2,10^3,10^4,10^5\}$. For this analysis, \loki was trained using only the second stage.

These results show that it is much more important to have more data in the second stage when compared to the first stage. The performance did not visibly improve over the data range considered for the first stage, indicating no benefit in training on a lot of existing links. On the other hand, the model performance improved drastically as the data size increased for the second stage, with no sign of plateauing. Based on these results, the optimal training schedule for an entity insertion model using our data seems to be a short first stage, followed by a second stage using as much data as possible.

\subsection{Training stages}

Table~\ref{tab:stage_analysis} shows the impact of different training strategies: (1) Warm start (only using the first stage), (2) Expansion (only using the second stage), and (3) Warm start + Expansion (using both stages), on the downstream entity insertion performance of \loki using data extracted from English Wikipedia.

\begin{table*}[!htb]
    \centering
    \caption{Comparison of the impact of different stages of the training pipeline on the downstream entity insertion performance. The categorization of entity insertion types into `Overall', `Missing', and `Present' is discussed in \S~\ref{subsec:setup}.}
    \label{tab:stage_analysis}
    \begin{threeparttable}        
        \begin{tabular}{c|ccc|ccc}
        \multirow{2}{*}{Training Stages} & \multicolumn{3}{c|}{Hits@1} & \multicolumn{3}{c}{MRR} \\
         & Overall & Present & Missing & Overall & Present & Missing \\\hline
        Warm start & 0.584 & \textbf{0.883} & 0.350 & 0.649 & \textbf{0.907} & 0.451 \\
        Expansion & 0.604 & 0.738 & 0.494$^{\dagger}$ & 0.689 & 0.801 & 0.603$^{\dagger}$ \\
        Warm start + Expansion & \textbf{0.672}$^{\dagger}$ & 0.877$^{\dagger}$ & \textbf{0.509} & \textbf{0.744}$^{\dagger}$ & 0.906$^{\dagger}$ & \textbf{0.617} \\\hline
        \end{tabular}
        \begin{tablenotes}
            \small \item[\dag] Indicates statistical significance ($p<0.05$) between the variant and the previous variant.
        \end{tablenotes}
    \end{threeparttable}
\end{table*}

\subsection{RoBERTa vs XLM-RoBERTa}

We found the scores obtained with RoBERTa on Simple English to be significantly higher than the scores achieved by the multilingual XLM-RoBERTa. In this analysis, we compare the performance of these two models on the full English dataset, with both models having been fine-tuned on the same English dataset. Table~\ref{tab:en_vs_multi} shows a statistically significant difference in the performance of RoBERTa and XLM-RoBERTa, with RoBERTa scoring higher in all entity insertion strategies, with gaps up to 25\%. We draw two conclusions from these results.

\begin{table*}[!htb]
    \centering
    \caption{Comparing the entity insertion performance of our model fine-tuned using the monolingual RoBERTa$_{\text{BASE}}$ and the multilingual XLM-RoBERTa$_{\text{BASE}}$ on the data extracted from English Wikipedia. The categorization of entity insertion types is discussed in \S~\ref{sec:data}.}
    \label{tab:en_vs_multi}
    \begin{threeparttable}        
        \resizebox{\textwidth}{!}{%
        \begin{tabular}{c|cccccccc}
        \multirow{2}{*}{Model} & \multicolumn{2}{c}{Text Present} & \multicolumn{2}{c}{Missing Mention} & \multicolumn{2}{c}{Missing Sentence} & \multicolumn{2}{c}{Missing Span} \\
         & Hits@1 & MRR & Hits@1 & MRR & Hits@1 & MRR & Hits@1 & MRR \\\hline
         RoBERTa$_{\text{BASE}}$ & \textbf{0.923}$^{\dagger}$ & \textbf{0.936}$^{\dagger}$ & \textbf{0.737}$^{\dagger}$ & \textbf{0.797}$^{\dagger}$ & \textbf{0.850}$^{\dagger}$ & \textbf{0.898}$^{\dagger}$ & \textbf{0.787}$^{\dagger}$ & \textbf{0.848}$^{\dagger}$ \\
         XLM-RoBERTa$_{\text{BASE}}$ & 0.863 & 0.892 & 0.543 & 0.630 & 0.595 & 0.662 & 0.697 & 0.615 \\\hline
        \end{tabular}
        }
        \begin{tablenotes}
            \small \item[\dag] Indicates statistical significance ($p<0.05$).
        \end{tablenotes}
    \end{threeparttable}
\end{table*}

In our ablations, we found that RoBERTa outperformed BERT and T5$^{\text{enc}}$ by a large margin, which leads us to select XLM-RoBERTa as the best candidate for the multilingual model to use in our experiments. However, the performance of RoBERTa does not seem to directly correlate with the performance of XLM-RoBERTa, as seen by the large drop in English when moving from RoBERTa to XLM-RoBERTa. This finding casts some doubt on the decision of the best multilingual model and opens the doors to models like multilingual BERT and mT5 \cite{mt5}. In the future, it would be interesting to consider other multilingual models and see if they can outperform XLM-RoBERTa.

As shown in \S~\ref{sec:multilingual_results}, XLM-RoBERTa fine-tuned on the multilingual dataset generally outperformed XLM-RoBERTa fine-tuned on a single language. However, the results in Table~\ref{tab:en_vs_multi} point to the fact that a model pre-trained on a single language (RoBERTa) outperforms a model pre-trained on multiple languages (XLM-RoBERTa). The dominance of the monolingual model is not surprising as a model pre-trained on a single language had a much smaller domain to learn than a multilingual model, and thus, might have been able to learn linguistic and structural patterns that the multilingual model failed to capture. So, for the languages where a pre-trained model does exist (for example, BERT for English, CamemBERT \cite{camembert} for French, HerBERT \cite{herbert} for Polish), that model may outperform the multilingual variant. However, it is unrealistic to assume that there can be a pre-trained model for each of the 300+ languages of Wikipedia. The multilingual model becomes essential for the languages for which there is no pre-trained model. As we saw in \S~\ref{sec:multilingual_results} and \S~\ref{sec:zero_shot}, the multilingual model is capable of transferring knowledge to unseen languages, which proves its potential for low-resource languages for which a full pre-trained model is not realistic.

\begin{table*}[!htb]
    \centering
    \caption{Comparing the entity insertion performance obtained for Simple English with different loss functions: pointwise \vs ranking loss. The categorization of entity insertion types into `Overall', `Missing', and `Present' is discussed in \S~\ref{subsec:setup}.}
    \begin{tabular}{c|ccc|ccc}
        \multirow{2}{*}{Method} & \multicolumn{3}{c|}{Hits@1} & \multicolumn{3}{c}{MRR} \\
        & Overall & Present & Missing & Overall & Present & Missing \\\hline
        Pointwise Loss & 0.641 & 0.891 & 0.477 & 0.712 & 0.922 & 0.574 \\
        Ranking Loss & \textbf{0.658} & \textbf{0.907} & \textbf{0.495} & \textbf{0.731} & \textbf{0.930} & \textbf{0.601} \\\hline
    \end{tabular}
    \label{tab:loss_comparison}
\end{table*}

\subsection{Single Encoder vs Triple Encoder}

In early iterations of our work, we explored a different model architecture. This architecture used the additional knowledge of the source article. Given the amount of text that needed to be encoded, and considering that most transformers have a limited number of tokens they can process, we chose to encode each of the three components separately. We had the following input representations:
\begin{itemize}
    \item Source Article: [CLS]<Src Title>[SEP]<Src Lead>
    \item Candidate: [CLS]<Src Section>[SEP]<Tgt Mention>[SEP]<Context>
    \item Target Title: [CLS]<Tgt Title>[SEP]<Tgt Lead>
\end{itemize}

Each of the components of the triplet was encoded independently, and then stacked together. Finally, an MLP capturing the interactions between the three embeddings was used to produce a relevance score.

The key intuition behind this architecture was to represent a link as a knowledge triplet \texttt{<src, text, tgt>}, and the overall architecture was supposed to predict whether the triplet was correct. However, we found that such an architecture decayed into a state where the target and source embeddings were independent of the input, always producing the same embedding. We believe that the model relied exclusively on the semantic knowledge contained in the list of target mentions to identify whether the entity should be inserted in the candidate text span, and the source and target article embeddings decayed into a global average optimum that maximized the performance of the MLP for the candidate embedding. Nevertheless, this meant that all the knowledge about the target entity contained in the target lead was being ignored.

To take advantage of the total available information, we moved to the architecture described in \S~\ref{sec:architecture}. We removed the source title and the source lead, driven by the token limit of the transformer architecture. We believed that this knowledge provided the least marginal gain from the three components of the triplet, at a cost of token space for the candidate and the target, as the source article knowledge only gave additional context to the candidate text span. 

We additionally moved to a single encoder for two reasons. First, the transformer architecture is more expressive than an MLP, and thus, it was better suited to capture the interactions between the candidate and the target. With only two knowledge sources instead of three, we felt we had sufficient token space for each source to capture enough semantic information for each input. Second, by relying on one single embedding, the embedding couldn't decay into a global average optimum which provided no information about the input, because the relevance score was entirely dependent on the representation power of that single embedding.

\subsection{Pointwise Loss vs Ranking Loss}

Table~\ref{tab:loss_comparison} shows how the choice of different loss functions (pointwise \vs ranking) impacts the downstream entity insertion performance of our models evaluated on the Simple English dataset.

\end{document}